\documentclass{article} 
\usepackage{times}
\usepackage{iclr2023_conference}

\usepackage{amsmath,amsfonts,bm}

\def\eqref#1{equation~\ref{#1}}

\def\1{\bm{1}}

\def\rvm{{\mathbf{m}}}

\def\rvo{{\mathbf{o}}}

\def\rvs{{\mathbf{s}}}

\def\rvx{{\mathbf{x}}}

\def\rvz{{\mathbf{z}}}

\def\mA{{\bm{A}}}

\def\mI{{\bm{I}}}

\DeclareMathAlphabet{\mathsfit}{\encodingdefault}{\sfdefault}{m}{sl}
\SetMathAlphabet{\mathsfit}{bold}{\encodingdefault}{\sfdefault}{bx}{n}

\newcommand{\R}{\mathbb{R}}

\newcommand{\softmax}{\mathrm{softmax}}

\DeclareMathOperator*{\argmin}{arg\,min}

\usepackage{graphicx}
\usepackage{times}
\usepackage{comment}
\usepackage{amsmath,amssymb,amsfonts,mathabx,mathbbol,bm}
\usepackage{acronym}
\usepackage{enumitem}
\PassOptionsToPackage{dvipsnames,table}{xcolor}
\usepackage[pagebackref,colorlinks,breaklinks,citecolor=gray]{hyperref}
\usepackage{url}
\usepackage{balance}
\usepackage{xspace}
\usepackage{setspace}
\usepackage[skip=3pt,font=small]{subcaption}
\usepackage[skip=3pt,font=small]{caption}
\usepackage{caption}
\usepackage[misc]{ifsym}
\usepackage[capitalise]{cleveref}
\usepackage{tabularx}
\usepackage{multirow,multirow,array,makecell}
\usepackage{overpic}
\usepackage{wrapfig}
\usepackage{pifont}
\usepackage{url}
\usepackage{booktabs,tabularx,colortbl}       
\usepackage[utf8]{inputenc} 
\usepackage[T1]{fontenc}    
\usepackage{nicefrac}       
\usepackage{microtype}      
\usepackage[noend,ruled, vlined]{algorithm2e}
\usepackage{algorithmic}
\usepackage{rotating}
\usepackage{adjustbox}
\usepackage{hvfloat}

\makeatletter
\DeclareRobustCommand\onedot{\futurelet\@let@token\@onedot}
\def\@onedot{\ifx\@let@token.\else.\null\fi\xspace}
\def\eg{\emph{e.g}\onedot} 

\def\ie{\emph{i.e}\onedot}

\def\etc{\emph{etc}\onedot} 

\def\wrt{w.r.t\onedot}

\makeatother

\newcommand{\N}{\mathcal{N}}

\makeatletter
\renewcommand{\paragraph}{%
  \@startsection{paragraph}{4}%
  {\z@}{0ex \@plus 0ex \@minus 0ex}{-1em}%
  {\hskip\parindent\normalfont\normalsize\bfseries}%
}
\makeatother

\crefname{algorithm}{Alg.}{Algs.}
\Crefname{algocf}{Algorithm}{Algorithms}
\crefname{section}{Sec.}{Secs.}
\Crefname{section}{Section}{Sections}
\crefname{table}{Tab.}{Tabs.}
\Crefname{table}{Table}{Tables}
\crefname{figure}{Fig.}{Fig.}
\Crefname{figure}{Figure}{Figure}

\graphicspath{{figure/}}

\acrodef{nlp}[NLP]{Natural Language Processing}
\acrodef{vos}[VOS]{Video Object Segmentation}
\acrodef{ucla}[UCLA]{University of California, Los Angeles}
\acrodef{pku}[PKU]{Peking University}
\acrodef{thu}[THU]{Tsinghua University}
\acrodef{bigai}[BIGAI]{Beijing Institute of General Artificial Intelligence}
\acrodef{model}[BO-QSA]{Bi-level Optimized Query Slot Attention}
\acrodef{qsa}[QSA]{Query Slot Attention}
\acrodef{gru}[GRU]{Gated Recurrent Unit}
\acrodef{ari}[ARI]{Adjusted Rand Index}
\acrodef{msc}[MSC]{Mean Segmentation Covering}
\acrodef{isa}[I-SA]{Implicit Slot-Attention}
\acrodef{miou}[mIoU]{mean Intersection over Union}
\acrodef{mse}[MSE]{Mean Squared Error}
\acrodef{dvae}[dVAE]{discrete VAE}
\acrodef{ste}[STE]{straight-through gradient estimator}
\newcommand{\sa}{\text{Slot-Attention}\xspace}
\newcommand{\sota}{\text{state-of-the-art}\xspace}
\newcommand{\model}{\text{Bi-level Optimized Query Slot Attention}\xspace}
\newcommand{\meta}{\text{meta-learning}\xspace}
\newcommand{\abbrev}{\text{BO-QSA}\xspace}
\newcommand{\shortprefix}{\text{BO-}}

\definecolor{scope}{RGB}{103,78,167}
\definecolor{semantic}{RGB}{230,145,56}
\definecolor{type}{RGB}{153,0,0}
\definecolor{revision}{RGB}{0,0,0}

\makeatletter
\renewcommand*{\@fnsymbol}[1]{\ensuremath{\ifcase#1\or *\or \dagger\or \ddagger\or
   \mathsection\or \mathparagraph\or \|\or **\or \dagger\dagger
   \or \ddagger\ddagger \else\@ctrerr\fi}}
\makeatother

\title{Improving Object-centric Learning with Query Optimization}

\author{
Baoxiong Jia$^{1,3\dag}$\thanks{Equal contribution. $^\dag$Work done during internship at BIGAI.}~~,~Yu Liu$^{2,3\dag*}$,~Siyuan Huang$^{3}$\\
\small $^1$UCLA~~$^2$Tsinghua University~~$^3$
\small National Key Laboratory of General Artificial Intelligence, BIGAI
}

\iclrfinalcopy 
\begin{document}
\maketitle

\begin{abstract}
The ability to decompose complex natural scenes into meaningful object-centric abstractions lies at the core of human perception and reasoning. In the recent culmination of unsupervised object-centric learning, the \sa
module has played an important role with its simple yet effective design and fostered many powerful variants. These methods, however, have been exceedingly difficult to train without supervision and are ambiguous in the notion of object, especially for complex natural scenes. 
In this paper, we propose to address these issues by investigating the potential of learnable queries as initializations for \sa learning, uniting it with efforts from existing attempts on improving \sa learning with bi-level optimization.
With simple code adjustments on \sa, our model, \model, achieves \sota results on 3 challenging synthetic and 7 complex real-world datasets in unsupervised image segmentation and reconstruction, outperforming previous baselines by a large margin. We provide thorough ablative studies to validate the necessity and effectiveness of our design. Additionally, our model exhibits great potential for concept binding and zero-shot learning. Our work is made publicly available at \url{https://bo-qsa.github.io}.
\end{abstract}

\section{Introduction}\label{sec:introduction}
Objects, and their interactions, are the foundations of human cognition~\citep{spelke2007core}. The endowment on making abstractions from perception and organizing them systematically empowers humans the ability to accomplish and generalize across a broad range of tasks, such as scene modeling~\citep{bear2020learning}, visual reasoning~\citep{yi2019clevrer}, and simulating interactions~\citep{bear2020learning}. The key to such success lies in the emergence of symbol-like mental representations of object concepts~\citep{whitehead1928symbolism}. However, important as it is, disentangling object-centric concepts from visual stimuli is an exceedingly difficult task to accomplish with limited supervision~\citep{greff2020binding} and requires proper inductive biases~\citep{scholkopf2021toward}. 

Motivated by the development of symbolic thought in human cognition, slot-based representations, instance~\citep{greff2017neural,greff2019iodine, locatello2020sa}, sequential~\citep{gregor2015draw, burgess2019monet,engelcke2021genesisv2,goyal2019recurrent}, or spatial~\citep{crawford2019spatially,lin2020space,jiang2019scalor}, have been the key inductive bias to recent advances in unsupervised object-centric learning. Among them, the \sa module has received tremendous focus given its simple yet effective design~\citep{locatello2020sa}. By leveraging the iterative attention mechanism, \sa learns to compete between slots for explaining parts of the input, exhibiting a soft-clustering effect on visual signals. It is later proven to be more memory and training efficient as a plug-and-play module for unsupervised object-centric learning~\citep{locatello2020sa} and fostered powerful variants in understanding images~\citep{singh2021slate,xu2022groupvit}, 3D scenes~\citep{yu2021unsupervised, sajjadi2022object} and videos~\citep{kipf2021conditional,elsayed2022savi++,singh2022simple}.

However, as revealed by recent studies, the \sa module comes with innate discrepancies for object-centric representation learning. First, with slots randomly initialized each time, the object-centric representations obtained by these models do not necessarily bind to object concepts~\citep{kipf2021conditional}. Intuitively, such randomness leads to undesired scenarios where slots with similar initializations compete for objects on different images. Such randomness challenges the iterative refinement procedure as it now needs to project sets of potentially similar representations to independent constituents of the input. As discovered by~\cite{chang2022object}, differentiating through such recurrences contributes to various training instabilities with growing spectral norm of \sa weights. This leads to the second and perhaps least desired property of \sa; it relies heavily on hyper-parameter tuning, including gradient clipping, learning rate warm-up, \etc, and further hurts the flexibility of Slot-Attention in adapting to broader applications with more complex signals.

To this end, we propose an extension of the \sa module, \ac{model}, to tackle the aforementioned problems. First, 
we follow the bi-level optimization framework proposed by~\cite{chang2022object} for easing the training difficulty in \sa.
More importantly, instead of sampling from a learnable Gaussian distribution, we propose to directly learn the slot initializations as queries.
With these learnable representations, we eliminate the ambiguous competitions between slots and provide a better chance for them to bind to specific object concepts. 
We improve the training of query-initialized \sa with a \ac{ste} by connecting our method with first-order approaches~\citep{finn2017model,nichol2018reptile,geng2021training} in solving bi-level optimization problems. The experimental results show that the proposed~\ac{model} can achieve \sota results on both synthetic and real-world image datasets with simple code adjustments to the original \sa module. 

With our model significantly outperforming previous methods in both synthetic and real domains, we provide thorough ablative studies demonstrating the effectiveness of our model design. We later show that our~\ac{model} possesses the potential of binding object concepts to slots. To validate this potential, we design zero-shot transfer learning experiments to show the generalization power of our model on unsupervised object-centric learning. As the experiments suggest (see~\cref{sec:exp}),  our model could potentially be a principle approach for unsupervised object-centric learning and serve as a general plug-and-play module for a broader range of modalities where variants of \sa prosper. We hope these efforts can help foster new insights in the field of object-centric learning.

\paragraph{Contributions} In summary, our main contributions are three-fold:
\begin{itemize}[leftmargin=*,noitemsep,nolistsep]
    \item We propose \ac{model}, a query-initialized \sa model that unites straight-through gradient updates to learnable queries with methods on improving \sa with bi-level optimization.
    \item We show that, with simple code adjustments on \sa, the proposed \ac{model} achieves \sota results on several challenging synthetic and real-world image benchmarks, outperforming previous methods by a large margin.
    \item We show the potential of our \ac{model} being a better approach to concept binding and learning generalizable representations with qualitative results and zero-shot transfer learning experiments.
\end{itemize}

\section{Preliminaries}\label{sec:prelim}
\subsection{Object-Centric Representation Learning with \sa}\label{sec:prelim:slot_attention}
\sa~\citep{locatello2020sa} takes a set of $N$ input feature vectors  $\rvx \in \R^{N \times D_{\text{input}}}$ and maps them to a set of $K$ output vectors (\ie, slots) $\rvs \in \R^{K \times D_{\text{slots}}}$. It leverages an iterative attention mechanism to first map inputs and slots to the same dimension $D$ with linear transformations $k(\cdot)$, $q(\cdot)$ and $v(\cdot)$ parameterized by $\phi^{\text{attn}}$. At each iteration, the slots compete to explain part of the visual input by computing the attention matrix $\mA$ with softmax function over slots and updating slots with the weighted average of visual values:
\begin{equation*}
    \begin{aligned}
    \tilde{\rvs} = f_{\phi^{\text{attn}}}(\rvs, \rvx) = \left(\frac{A_{i,j}}{\sum_{l=1}^{N}A_{l, j}}\right)^{\top}\cdot v(\rvx) && \text{where} && \mA = \softmax\left(\frac{k(\rvx)\cdot q(\rvs)^{\top}}{\sqrt{D}}\right) \in \R^{N\times K}.
    \end{aligned}
\end{equation*}
The slots are initialized from a learnable Gaussian distribution with mean $\bm{\mu}$ and variance $\bm{\sigma}$. They are refined iteratively within the \sa module by passing the updates into a \ac{gru}~\citep{cho2014learning} and MLP parameterized by $\phi^{\text{update}}$ for $T$ iterations:
\begin{equation}
    \begin{aligned}
    \rvs^{(t+1)} = h_{\phi^\text{update}}(\rvs^{(t)}, \tilde{\rvs}^{(t)}), && \rvs^{0}\sim \N(\bm{\mu}, \text{diag}(\bm{\sigma})), && \hat{\rvs} = \rvs^{(T)}.
    \end{aligned}
    \label{eq:iterative_refine}
\end{equation}
The final prediction $\hat{\rvs}$ can be treated as the learned object-centric representation \wrt to input features $\rvx$. In the image domain, we take as input a set of images $\mI$ and encode them with $f_{\phi^{\text{enc}}}$ to obtain features $\rvx \in \R^{HW\times D_{\text{input}}}$. After obtaining $\hat{\rvs}$ through the iterative refinement procedure with $h_{\phi^{\text{update}}}$, images could be decoded from these object-centric representations with a mixture-based decoder or autoregressive transformer-based decoder. We refer the readers to~\cref{app:model:decoder} for details on different decoder designs and their ways of visualizing learned object concepts.

\subsection{Improving \sa with Bi-level Optimization}\label{sec:prelim:meta_learning}
The problem of bi-level optimization embeds the optimization of an inner objective within the outer objective. Normally, a bi-level optimization problem can be formulated as:
\begin{equation}
    \begin{aligned}
    \min_{\theta, \phi} f(\theta, \phi) && s.t. && \theta \in \argmin_{\theta'}g(\theta', \phi),
    \end{aligned}
    \label{eq:bo_supp}
\end{equation}
where we call $f(\theta,\phi)$ the outer objective function and $g(\theta, \phi)$ the inner objective function. To jointly optimize both objectives \wrt parameters $\theta$ and $\phi$, a straightforward approach to solving~\cref{eq:bo_supp} is to represent the inner solution of $\theta$ as a function of $\phi$, \ie, $\theta^*(\phi) = \argmin_{\theta'}g(\theta', \phi)$.
Then we can optimize the outer objective with gradient descent by approximating $\nabla_{\phi}f(\theta^*(\phi), \phi)$ as a function of $\phi$. 
When the inner optimization objective could be solved by a fixed point iteration $\theta = F_{\phi}(\theta)$~\citep{amos2017optnet,bai2019deep}, the bi-level optimization problem could be solved by
\begin{equation}
    \frac{\partial f(\theta^{*}(\phi), \phi)}{\partial {\phi}} = \frac{\partial f(\theta^*(\phi), \phi)}{\partial\theta^*}\cdot\sum_{i=0}^{\infty}\left(\frac{\partial F_{\phi}(\theta^*)}{\partial \theta^*}\right)^{i}\cdot\frac{\partial F_{\phi}(\theta^*)}{\partial \phi}.
    \label{eq:I-SA}
\end{equation}
For efficiency concerns, recent methods often use the first-order approximation of the infinite Neumann's series~\citep{shaban2019truncated,geng2021training} for updating $\phi$. Given that \sa is, in essence, an iterative refinement method that falls into the same framework,~\cite{chang2022object} adapted this technique to improve \sa training and obtained significant improvement both in model performance and training stability. We provide more discussions on this in~\cref{sec:model:meta_learning} and also other bi-level optimization methods for approximating $\nabla_{\phi}f(\theta^*(\phi), \phi)$ in~\cref{app:model:optimization}.

\section{Method}\label{sec:model}
\subsection{Query Slot Attention}\label{sec:model:qsa}
As mentioned in~\cref{sec:introduction}, the \sa module adopts a random initialization of slots and conducts iterative refinement to obtain object-centric representations $\hat{\rvs}$ as in~\cref{eq:iterative_refine}. However, as argued by~\cite{kipf2021conditional}, such random initializations provide no hint on the notion of object and no means for controllably probing concepts from the model. As shown by~\citet{chang2022object}, this random initialization plays a minimal role and could be detached from training. This indicates that the estimation of $\hat{\rvs}$ relies heavily on the task-specific iterative refining of slots over data, leaving a limited possibility for slots to bind to specific concepts and be leveraged as generalizable representations. 

To address this issue, we focus on the \ac{qsa}, which initializes the slots in the \sa module with learnable queries $\rvs_0 = \phi^\text{init}$.
Such a design is motivated by the success of recent query-based networks~\citep{van2017neural,jaegle2021perceiver}.
It facilitates an object-centric model to learn general symbolic-like representations that could be quickly adapted by refining over task-specific requirements, as discussed in \cref{sec:introduction} and ~\cite{kipf2021conditional}. 
Meanwhile, in contrast to the use of learnable queries in other encoder-decoder structures (\eg ~\ac{dvae}), the slot initializations $\rvs_0$ are not necessarily required to encode image features since they were designed for separating them. This resembles recent discoveries in query networks~\citep{carion2020end, yang2021self} where queries could be generalizable probes for input properties. Despite the good properties and potentials~\ac{qsa} presents, it is shown detrimental to initialize slots independently in \sa under unsupervised settings~\citep{locatello2020sa}.

\subsection{Rethinking Bi-level Optimization Methods for Query Slot Attention}\label{sec:model:meta_learning}
To improve the learning of \ac{qsa}, we rewind to the idea of improving the learning of the vanilla \sa module with bi-level optimization~\citep{chang2022object}.
Under this formulation, \sa could be treated as solving the following objectives:
\begin{equation}
    \begin{aligned}
    \min_{\rvs, \Phi} \sum_{i=1}^{M}\mathcal{L}(\rvx_{i}, \rvs_{i}, \Phi)
    && s.t. && \rvs_{i}^* = \argmin_{\rvs} \mathcal{L}_{\text{cluster}}(\rvx_{i}, \rvs, \Phi),
    \end{aligned}
    \label{eq:sa_meta}
\end{equation}
where $\rvx_{i}$ and $\rvs_{i}$ denote the input feature from the $i$-th image and its corresponding slots, and $\Phi =\{\phi^{\text{init}}, \phi^{\text{attn}}, \phi^{\text{update}}\}$ denotes parameters for assigning input features $\rvx$ to different slots. Under this setting, the outer objective $\mathcal{L}$ is usually a reconstruction objective and the inner objective could be viewed as a soft-clustering objective~\citep{locatello2020sa}.
Next, the inner objective is solved by iterative refinement, which could be formulated as solving for fixed-points~\citep{chang2022object} of
\begin{equation}
    \begin{aligned}
        \rvs = h_{\phi^{\text{update}}}(\rvs, \tilde{\rvs}) = h_{\phi^{\text{update}}}(\rvs, f_{\phi^{\text{attn}}}(\rvs, \rvx)) = F_{\Phi}(\rvs, \rvx),
    \end{aligned}
    \label{eq:fixed_point}
\end{equation}
where $F_{\Phi}(\cdot,\cdot)$ is an fixed-point operation. 
As introduced by~\cite{chang2022object} in ~\ac{isa}, with~\cref{eq:I-SA}, the instabilities through the iterative updates could be avoided by detaching gradients, treating slots in the final iteration as an approximation of $\rvs_{i}^{*}$, and computing first-order gradient approximations for updating $\Phi$ with $\rvs_{i}^*$. However, we demonstrate in~\cref{tab:ablation} that this design is only beneficial for randomly initialized slots and detrimental for query-initialized \sa architectures since it relies heavily on the good approximation of the solution to the inner objective. With no randomness in slot initializations or gradient during training, starting from a fixed set of initialization points puts challenges on the learning of \sa update $F_{\Phi}$ as it will be difficult to provide a good approximation of $s_{i}^{*}$ with only a fixed number of iterations (see in~\cref{app:exp:iqsa}). This urges the need for information flow to the slot initialization queries.

\subsection{\model}\label{sec:model:boqsa}

\begin{wrapfigure}[12]{c}{0.5\linewidth}
\raisebox{0pt}[\dimexpr\height-1.3\baselineskip\relax]{\begin{minipage}[r]{\linewidth}
\resizebox{\linewidth}{!}{
\small
\begin{algorithm}[H]
    \caption{\ac{model}}
    \label{alg:model}
    \DontPrintSemicolon
    \SetKwInOut{module}{Modules}
    \KwIn{input features \texttt{input}, learnable queries \texttt{init}, number of iterations $T$}
    \KwOut{object-centric representation \texttt{slots}}
    \module{stop gradient module SG($\cdot$), slot attention module SA($\cdot$, $\cdot$)}
    \texttt{slots} = \texttt{init}\\
    \For{$t = 1, \cdots, T$}{
        \texttt{slots} = SA(\texttt{slots}, \texttt{inputs})
    }
    \texttt{slots} = SG(\texttt{slots}) + \texttt{init} - SG(\texttt{init}) \\
    \texttt{slots} = SA(\texttt{slots}, \texttt{inputs}) \\
    \Return{\texttt{slots}}
\end{algorithm}
}
\end{minipage}
}
\end{wrapfigure}
We propose \ac{model} to address the learning problem of~\ac{qsa}. As shown in~\cref{alg:model}, we initialize slots with learnable queries in~\ac{model} and perform $T$ steps of \sa update to obtain an approximation of $\rvs_{i}^*$. These near-optimal solutions of the inner objective are passed into one additional \sa step where gradients to all previous iterations are detached. In contrary to~\ac{isa}, we use a \ac{ste}~\citep{bengio2013estimating,van2017neural} to backpropagate gradients and also to slot initialization queries. Such designs help find good starting points for the inner optimization problem on clustering, alleviating the problem of bi-level optimization with \ac{qsa} mentioned in~\cref{sec:model:meta_learning}. 
Similar to~\ac{dvae}, the \ac{ste} adds bias to the gradient of the initialization queries. However, since these learnable queries are meant for disentangling image features, they do not have to maintain information about the approximated $\rvs^*$. Such bias could lead to learned queries which are better pivots for separating different image features, similar to anchors, or filter queries learned for different tasks~\citep{carion2020end, zhang2021temporal}. Note that we do not add constraints on the consistency between $\rvs_0$ and $\hat{\rvs}$ (\eg $||sg(\hat{\rvs}) - \rvs_0||^2$) as done in~\ac{dvae} since we find such constraints lead to a mean-representation of datasets that forbids better concept binding (see in~\cref{app:exp:dvae}).
As shown in~\cref{tab:ablation} and~\cref{fig:tsne}, our learned slot initialization queries do fulfill this goal by providing a more separable initialization space and can significantly facilitate model learning.

\section{Related Work}\label{sec:related_work}
\paragraph{Unsupervised Object-Centric Learning} Our work falls into the recent line of research on unsupervised object-centric learning on images~\citep{greff2016tagger,eslami2016attend,greff2017neural,greff2019iodine,burgess2019monet,crawford2019spatially,engelcke2019genesis,lin2020space,bear2020learning,locatello2020sa, zoran2021parts}. A thorough review and discussion on this type of method can be found in~\cite{greff2020binding}. One critical issue of these methods is on handling complex natural scenes.~\cite{singh2021slate,lamb2021transformers} leverages a transformer-based decoder with \sa for addressing this problem. Similar attempts have also been made by exploiting self-supervised contrastive learning~\citep{choudhury2021unsupervised, caron2021emerging,wang2022self,henaff2022object} and energy-based models~\citep{du2021unsupervised, yu2021unsupervised}. Our work builds upon \sa by extending it with learnable queries and a novel optimization method for learning. Our compelling experimental suggests our model could potentially serve as a general plug-and-play module for a wider range of modalities where variants of \sa prosper~\citep{kipf2021conditional,elsayed2022savi++,singh2022simple, yu2021unsupervised,sajjadi2022object,sajjadi2022scene}.

\paragraph{Query Networks} Sets of latent queries are commonly used in neural networks. These methods leverage permutation equivariant network modules (\eg GNNs~\citep{scarselli2008graph} and attention modules~\citep{vaswani2017attention}) in model design for solving set-related tasks such as clustering~\citep{lee2019set}, outlier detection~\citep{zaheer2017deep, zhang2019deep}, \etc. These learned latent queries have been shown to have good potential as features for tasks like contrastive learning~\citep{caron2020unsupervised}, object detection~\citep{carion2020end}, and data compression~\citep{jaegle2021perceiverio,jaegle2021perceiver}. In contrast to the recent success of query networks in supervised or weakly-supervised learning~\citep{carion2020end,zhang2021temporal,kipf2021conditional,elsayed2022savi++,xu2022groupvit},~\cite{locatello2020sa} demonstrates the detrimental effect of using independently initialized slots in \sa learning. However, we show that our~\ac{model} method successfully overcomes this issue and generalizes the success of query networks to the domain of unsupervised object-centric learning.

\paragraph{Bi-level Optimization} Our work is closely related to bi-level optimization methods with iterative fixed update rules for solving the inner objective. Specifically, methods are designed with implicit differentiation~\citep{amos2017optnet,bai2019deep} to stabilize the iterative update procedure. Similar formulations are also found when combined with \meta where~\cite{madan2021fast} train queries through recurrence in a \meta fashion and~\cite{rajeswaran2019meta} provides a unified view of the optimization problem with implicit gradients.
Concurrent work from~\cite{chang2022object} formulate the \sa learning from an implicit gradient perspective with gradient stopping derived from first-order hyper-gradient methods~\citep{geng2021training}. However, they ignore the important role of slot initializations in generalization and concept binding. 
As our experiments suggest, such gradient-stopping methods do not guarantee superior performance compared to the original~\sa. We leave the details to~\cref{sec:exp:ablation} for an in-depth discussion.

\section{Experiments}\label{sec:exp}
In this section, we aim to address the following questions with our experimental results:
\begin{itemize}[leftmargin=*,noitemsep,nolistsep]
    \item How good is our proposed~\ac{model} on both synthetic and complex natural scenes?
    \item How important is the query and the optimization method in~\ac{model}?
    \item Does~\ac{model} possess the potential for concept binding and zero-shot transfer?
\end{itemize}

We provide details in the following sections with thorough comparative and ablative experiments and leave the details on model implementation and hyperparameter selection to~\cref{app:model:implement}. Here we clarify the datasets and metrics selected for evaluating our model on each domain:
\paragraph{Synthetic Domain} For the synthetic domain, we select three well-established challenging multi-object datasets Shapestacks~\citep{groth2018shapestacks}, ObjectsRoom~\citep{multiobjectdatasets19}, and CLEVRTEX for evaluating our~\ac{model} model. Specifically, we consider three metrics to evaluate the quality of object segmentation and reconstruction. ~\ac{ari}~\citep{hubert1985comparing} and~\ac{msc}~\citep{engelcke2019genesis} for segmentation and~\ac{mse} for reconstruction. Following the evaluation setting of recent works, we report the first two segmentation metrics over foreground objects (ARI-FG and MSC-FG). Additionally, we conduct extra experiments on more datasets and leave the discussion to~\cref{app:exp:results}.

\paragraph{Real-world Images} For the real image domain, we use two tasks (1) unsupervised foreground extraction and (2) unsupervised multi-object segmentation for evaluating our method. Specifically, we select Stanford Dogs~\citep{khosla2011novel}, Stanford Cars~\citep{krause20133d}, CUB200 Birds~\citep{welinder2010caltech}, and Flowers~\citep{nilsback2010delving} as our benchmarking datasets for foreground extraction and YCB~\citep{calli2017yale}, ScanNet~\citep{dai2017scannet}, COCO~\citep{lin2014microsoft} proposed by~\cite{yang2022promising} for multi-object segmentation.
We use \ac{miou} and Dice as metrics for evaluating the quality of foreground extraction and use the evaluation metrics adopted by~\cite{yang2022promising} for multi-object segmentation.
\subsection{Object Discovery on Synthetic Datasets}\label{sec:exp:synthetic}
\paragraph{Experimental Setup} We explore our proposed~\ac{model} with two types of decoder designs, mixture-based and transformer-based, as discussed in~\cref{sec:prelim:slot_attention} and~\cref{app:model:decoder}. We follow the decoder architecture in \sa~\citep{locatello2020sa} for mixture-based decoders and SLATE~\citep{singh2021slate} for transformer-based decoders. For both types of models, we use the \sa module with a CNN image encoder and initialize slots with learnable embeddings. 

\begin{table}[t!]
\centering
\caption{Multi-object segmentation results on ShapeStacks and ObjectsRoom. We report ARI-FG and MSC-FG of all models with (mean $\pm$ variance) across 3 experiment trials. We visualize the best results in bold.}
\label{tab:syn_seg}
\resizebox{0.83\linewidth}{!}{
\tiny
\begin{tabular}{ccccc}
    \toprule
    \multirow{2}[2]{*}{Model} & \multicolumn{2}{c}{ShapeStacks} & \multicolumn{2}{c}{ObjectsRoom}\\
    \cmidrule(lr){2-3}\cmidrule(lr){4-5}
    &$\uparrow$ ARI-FG &$\uparrow$ MSC-FG &$\uparrow$ ARI-FG &$\uparrow$ MSC-FG \\
    \midrule
    MONet-G~\citep{burgess2019monet} &0.70$\pm$0.04 &0.57$\pm$0.12 & 0.54$\pm$0.00 &0.33$\pm$0.01 \\
    GENESIS~\citep{engelcke2019genesis} & 0.70$\pm$0.05 & 0.67$\pm$0.02 & 0.63$\pm$0.03 & 0.53$\pm$0.07 \\
    \sa~\citep{locatello2020sa} & 0.76$\pm$0.01 & 0.70$\pm$0.05 & 0.79$\pm$0.02 & 0.64$\pm$0.13\\
    GENSIS-V2~\citep{engelcke2021genesisv2} & 0.81$\pm$0.01 & 0.67$\pm$0.01 &0.86$\pm$0.01 & 0.59$\pm$0.01\\
    SLATE~\citep{singh2021slate} &0.65$\pm$0.03 & 0.63$\pm$0.05 & 0.57$\pm$0.03 & 0.30$\pm$0.03 \\
    I-SA~\citep{chang2022object} &0.90$\pm$0.02 & 0.85$\pm$0.03 & 0.85$\pm$0.01 & 0.76$\pm$0.04 \\
    \midrule
    Ours (transformer) &0.68$\pm$0.02 &0.70$\pm$0.02 &0.68$\pm$0.03 &0.72$\pm$0.03 \\
    Ours (mixture) &\textbf{0.93$\pm$0.01} &\textbf{0.89$\pm$0.00}  &\textbf{0.87$\pm$0.03} &\textbf{0.80$\pm$0.02} \\
    \bottomrule
\end{tabular}
}
\end{table}

\begin{table}[t!]
\centering
\caption{Multi-object segmentation results on CLEVRTEX. We report ARI-FG (\%) and MSE of all models in the form of (mean $\pm$ variance) across 3 experiment trials. We visualize the best results in bold.}
\label{tab:syn_add_clevrtex}
\resizebox{0.83\linewidth}{!}{
\small
\begin{tabular}{ccccccc}
    \toprule
    \multirow{2}[2]{*}{Model} & \multicolumn{2}{c}{CLEVRTEX-FULL} & \multicolumn{2}{c}{CLEVRTEX-OOD} & \multicolumn{2}{c}{CLEVRTEX-CAMO}\\
    \cmidrule(lr){2-3}\cmidrule(lr){4-5}\cmidrule(lr){6-7}
    & $\uparrow$ ARI-FG & $\downarrow$ MSE & $\uparrow$ ARI-FG & $\downarrow$ MSE & $\uparrow$ ARI-FG & $\downarrow$ MSE\\
    \midrule
    MONet~\citep{burgess2019monet} & 19.78$\pm$1.02 &\textbf{146$\pm$7} & 37.29$\pm$1.04 & 409$\pm$3 & 31.52$\pm$0.87 & 265$\pm$1\\
    \sa~\citep{locatello2020sa} & 62.40$\pm$2.33 & 254$\pm$8 & 58.45$\pm$1.87 & 487$\pm$16 & 57.54$\pm$1.01 & \textbf{215$\pm$7}\\
    GENSIS-V2~\citep{engelcke2021genesisv2} & 31.19$\pm$12.41 & 315$\pm$106 & 29.04$\pm$11.23 &539$\pm$147 & 29.60$\pm$12.84&278$\pm$75\\
    DTI~\citep{monnier2021unsupervised} & 79.90$\pm$1.37 &438$\pm$22 & 73.67$\pm$0.98& 590$\pm$4& \textbf{72.90$\pm$1.89} & 377$\pm$17\\
    I-SA~\citep{chang2022object} & 78.96$\pm$3.88 &280$\pm$8 & 83.71$\pm$0.88& \textbf{241$\pm$4}& 57.20$\pm$13.28 & 295$\pm$30\\
    \midrule
    Ours (mixture) &\textbf{80.47$\pm$2.49} &268$\pm$2 & \textbf{86.50$\pm$0.19} & 265$\pm$25 & 63.71$\pm$6.11 & 280$\pm$7 \\
    \bottomrule
\end{tabular}
}
\end{table}

\paragraph{Results} 
We report multi-object segmentation results on synthetic datasets in~\cref{tab:syn_seg} and visualize qualitative results in~\cref{fig:result_vis}. As shown in~\cref{tab:syn_seg}, our~\ac{model} achieves the \sota results with large improvements over previous object-centric learning methods on all metrics in ShapeStacks and ObjectsRoom. We also observe more stable model performance, \ie smaller variances in results, across different trials of experiments. Our model with mixture-based decoders obtains the best overall performance on all datasets.
More specifically, our mixture-based~\ac{model} significantly outperforms the vanilla \sa model ($\sim$15\%) with minimal architectural differences. This validates the importance of the learnable queries and our optimization method. We will continue this discussion in~\cref{sec:exp:ablation}. As shown in~\cref{tab:syn_add_clevrtex}, our model also achieves \sota results on the unsupervised object segmentation task in CLEVRTEX with consistent improvement over \sa on the CAMO and OOD generalization split. Interestingly, our model (1) shows larger reconstruction errors, (2) generalizes well in out-of-distribution scenarios, and (3) shows marginal improvement in camouflaged images. We attribute (1) and (3) to the simple architecture of encoders/decoders currently adopted and provide insights on (2) in~\cref{sec:exp:transfer}.

\begin{wrapfigure}[9]{c}{0.5\linewidth}
\raisebox{0pt}[\dimexpr\height-1.4\baselineskip\relax]{\begin{minipage}{\linewidth}
\captionof{table}{Reconstruction results on ShapeStacks and ObjectsRoom 
(MSE$\downarrow$). We compare mixture-based and transformer-based decoder designs.}
\label{tab:syn_mse}
\resizebox{\linewidth}{!}{
\begin{tabular}{ccc}
\toprule
Model & ShapeStacks & ObjectsRoom \\
\midrule
\sa (mixture) & 80.8 & 20.4 \\
ours (mixture) & \textbf{72.0} & \textbf{8.1} \\
\midrule
SLATE (transformer) & 52.3 & 16.3 \\
ours (transformer) & \textbf{49.3} & \textbf{14.7} \\
\bottomrule
\end{tabular}
}
\end{minipage}}
\end{wrapfigure}
\paragraph{Mixture-based vs. Transformer-based Decoder} We observe inferior segmentation but 
superior reconstruction performance of 
transformer-based variants of \sa on synthetic datasets.
Specifically, we compare 
the~\ac{mse} of models on ShapeStacks and ObjectsRoom.
As shown in~\cref{tab:syn_mse},
transformer-based methods provide 
better reconstruction results. We attribute the low segmentation performance to mask prediction in these methods, which relies on the attention matrix computed over input features. 
This leads to coarse object masks as a result of image tokenization. 
Nonetheless, we observe consistent improvement by applying our slot encoder to both mixture and transformer decoders.

\subsection{Object Discovery on Real Datasets}\label{sec:exp:real}
\paragraph{Experimental Setup} For real-world experiments, we use the same slot encoder design used in~\cref{sec:exp:synthetic} with a 4-layer CNN image encoder and initialize slots with learnable queries. For unsupervised foreground extraction, we follow~\cite{yu2021drc} and report the best model performance on all datasets. During the evaluation, we select the slot's mask prediction that has a maximum intersection with the ground-truth foreground mask as our predicted foreground. For unsupervised multi-object segmentation, we follow~\cite{yang2022promising} and report the models' performance on all datasets across trials with different random seeds.

\begin{table}[t!]
    \centering
    \caption{Unsupervised multi-object segmentation results on YCB, ScanNet, and COCO variant proposed by~\cite{yang2022promising}. We use the same evaluation metrics as in~\cite{yang2022promising} and report all models' results with (mean (variance)) over 3 experiment trials. We visualize the best results in bold.}
    \label{tab:real_multi_seg}
    \resizebox{\linewidth}{!}{
    \begin{tabular}{cccc}
    \toprule
    \multirow{2}[2]{*}{Model} & YCB & ScanNet & COCO \\
     & (AP / PQ / Pre / Rec) $\uparrow$ & (AP / PQ / Pre / Rec) $\uparrow$ & (AP / PQ / Pre / Rec) $\uparrow$ \\
     \midrule
     AIR~\citep{eslami2016attend}& 0.0 (0.1) /0.6 (0.3) / 1.1 (0.4) / 0.8 (0.2) & 2.7 (1.4) / 6.3 (1.7) / 15.6 (2.8) / 7.3 (1.6) & 2.7 (0.1) / 6.7 (0.5) / 14.3 (2.6) / 8.6 (0.8) \\
     MONet~\citep{burgess2019monet} & 3.1 (1.6) / 7.0 (2.6) / 9.8 (3.6) / 1.2 (0.8) & 24.8 (1.6) / 24.6 (1.6) / 31.0 (1.6) / 40.7 (1.8) & 11.8 (2.0) / 12.5 (1.1) / 16.1 (0.9) / 21.9 (1.7) \\
     IODINE~\citep{greff2019iodine} & 1.8 (0.2) / 3.9 (1.3) / 6.2 (2.0) / 7.3 (1.9) & 10.1 (2.9) / 13.7 (2.7) / 18.6 (4.2) / 24.4 (3.8) & 4.0 (1.2) / 6.3 (1.2) / 9.9 (1.8) / 10.8 (2.0) \\
     \sa~\citep{locatello2020sa} & 9.2 (0.4) / 13.5 (0.9) / 20.0 (1.3) / 26.2 (6.8) & 5.7 (0.3) / 9.0 (1.5) / 12.4 (2.5) / 18.3 (2.7) & 0.8 (0.3) / 3.5 (1.2) / 5.3 (1.7) / 7.3 (2.2) \\
     I-SA~\citep{chang2022object} & 31.5 (15.2) / 25.6 (9.0) / 38.1 (12.5) / 40.2 (11.9) & 21.4 (6.8) / 23.4 (1.5) / 29.1 (7.8) / 34.5 (7.0) & 12.8 (4.8) / 13.7 (4.5) / 20.4 (6.0) / 24.6 (7.3) \\
     \midrule
     Ours (transformer) & \textbf{48.0 (1.8)} / \textbf{34.8 (1.3)} / \textbf{50.8 (1.1)} / \textbf{53.6 (0.7)} & \textbf{28.5 (2.4)} / \textbf{26.4 (2.0)} / \textbf{37.3 (2.0)} / \textbf{42.4 (1.9)} & \textbf{17.8 (0.6)} / \textbf{17.6 (0.6)} / \textbf{25.3 (0.6)} / \textbf{30.6 (0.9)} \\
     \bottomrule
    \end{tabular}
    }
\end{table}
\begin{table}[t!]
\centering
\caption{Unsupervised foreground extraction results on CUB200 Birds (Birds), Stanford Dogs (Dogs), Stanford Cars (Cars), and Caltech Flowers (Flowers). We visualize the best results in bold.} \label{tab:real_seg}
\resizebox{0.83\linewidth}{!}{
\small
\begin{tabular}{ccccccccc}
\toprule
\multirow{2}[2]{*}{Model} & \multicolumn{2}{c}{Birds} & \multicolumn{2}{c}{Dogs} & \multicolumn{2}{c}{Cars} & \multicolumn{2}{c}{Flowers}\\
\cmidrule(lr){2-3}\cmidrule(lr){4-5}\cmidrule(lr){6-7}\cmidrule(lr){8-9}
&$\uparrow$ IoU &$\uparrow$ Dice &$\uparrow$ IoU &$\uparrow$ Dice &$\uparrow$ IoU &$\uparrow$ Dice
&$\uparrow$ IoU &$\uparrow$ Dice\\
\midrule
ReDO~\citep{chen2019unsupervised} & 46.5 & 60.2 & 55.7 & 70.3 & 52.5 & 68.6 & 76.4 & -\\
IODINE~\citep{greff2019iodine} & 30.9 & 44.6 & 54.4 & 67.0 & 51.7 & 67.3 & - & -\\
OneGAN~\citep{benny2020onegan} &55.5 &69.2 &71.0 &81.7 &71.2 &82.6 & - & -\\
\sa~\citep{locatello2020sa} &35.6 &51.5 &39.6 &55.3 &41.3 &58.3 &30.8 &45.9\\
\cite{voynov2020big}  & 68.3 &- &- &- &- &- & 54.0 & -\\
DRC~\citep{yu2021drc} &56.4 &70.9 &71.7 &83.2 &72.4 &83.7 & - & -\\
\cite{melas2021finding} & 66.4 &- &- &- &- &- & 54.1 & -\\
SLATE~\citep{singh2021slate} &36.1 &51.0 &62.3 &76.3 &75.5 &85.9 & 68.1 & 79.1 \\
I-SA~\citep{chang2022object} &63.7 &72.7 &80.6 &89.1 &85.9 &92.3 & 75.0 & 83.9 \\
\midrule
Ours (mixture) &25.1 &39.2 &36.8 &53.6 &69.1 &81.5 & 36.1 & 51.6 \\
Ours (transformer) &\textbf{71.0} &\textbf{82.6} &\textbf{82.5} &\textbf{90.3} &\textbf{87.5} &\textbf{93.2} &\textbf{78.4} &\textbf{86.1}\\
\bottomrule
\end{tabular}
}
\end{table}
\begin{figure}[t!]
    \centering
    \resizebox{\linewidth}{!}{\includegraphics[width=\linewidth]{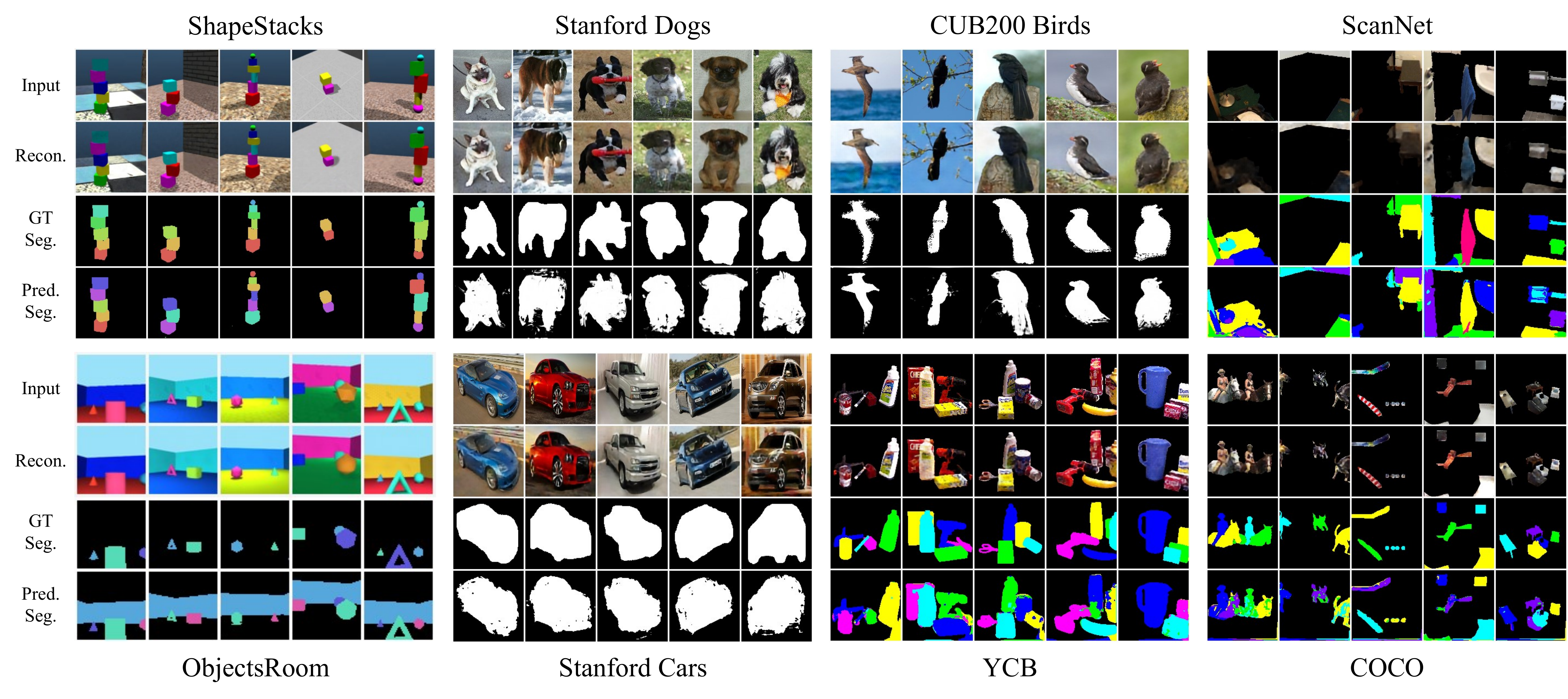}}
    \caption{Visualization of our predicted segmentation and reconstruction results on synthetic and real images. We color the predicted mask that has a maximum intersection with the ground-truth background in black.}
    \label{fig:result_vis}
\end{figure}

\begin{wrapfigure}[10]{c}{0.4\linewidth}
\raisebox{0pt}[\dimexpr\height-2.6\baselineskip\relax]{\begin{minipage}{\linewidth}
\captionof{table}{Unsupervised segmentation results on Birds (mIoU$\uparrow$). *Contrastive learning methods are pre-trained on ImageNet and segment with K-means clustering.}
\label{tab:real_contrast}
\tiny
\resizebox{\linewidth}{!}{
\begin{tabular}{cc}
\toprule
Model & Birds\\
\midrule
MoCo v2~\citep{chen2020improved} & 63.5\\
BYOL~\citep{grill2020bootstrap} & 56.1\\
R2O~\citep{gokul2022refine} & \textbf{71.2}\\
\midrule
ours (\ac{model}+transformer) & \textbf{71.0}\\
\bottomrule
\end{tabular}
}
\end{minipage}}
\end{wrapfigure}
\paragraph{Results} We show quantitative experimental results in
~\cref{tab:real_seg} and~\cref{tab:real_multi_seg}. We also visualize qualitative results in~\cref{fig:result_vis}. For multi-object segmentation, as shown in~\cref{tab:real_multi_seg}, our model outperforms existing object-centric learning baselines by a large margin, especially on the YCB dataset where the segmented objects have clear semantic meanings.
For foreground extraction, as shown in~\cref{tab:real_seg}, our method significantly outperforms all existing baselines on the task of foreground extraction, achieving new \sota on all datasets. We recognize the discrepancy of mixture-based decoders in both \sa and our mixture-based design in modeling real-world images, reflecting similar discoveries from recent works~\citep{singh2021slate} that 
mixture-based 
decoder struggles 
in modeling real-world images.
On the other hand, 
our transformer-based model shows significant
improvements over the vanilla version.
Notably, our method outperforms a broad range of models, including GAN-based generative models (\ie OneGAN,~\cite{voynov2020big}), and large-scale pre-trained contrastive methods (\ie MoCo-v2, BYOL, R2O). As shown in~\cref{tab:real_contrast}, our method achieves comparable results with 
\sota self-supervised contrastive learning methods without large-scale pre-training and data augmentation. This result sheds light on the potential of object-centric learning as a pre-training task for learning general visual representations.
\subsection{Ablative Studies}\label{sec:exp:ablation}
\begin{table}[t!]
\begin{minipage}{0.52\linewidth}

\centering
\caption{Ablative experiments on slot initialization and optimization methods. We visualize the best results in bold and underline the second-best results. (*Note that SA represents \sa with our encoder-decoder design and is different from the original one reported in~\cref{tab:real_seg}.)} \label{tab:ablation}
\resizebox{\linewidth}{!}{
\begin{tabular}{ccccc}
\toprule
\multirow{2}[2]{*}{Method} & \multicolumn{2}{c}{Dogs} &  \multicolumn{2}{c}{ShapeStacks} \\
\cmidrule(lr){2-3}\cmidrule(lr){4-5}
& $\uparrow$ IoU &$\uparrow$ Dice & $\uparrow$ ARI-FG(\%) &$\uparrow$ MSC-FG(\%) \\
\midrule
SA* & 71.0 & 81.9 & 86.7 & \underline{84.8} \\
I-SA & 80.8 & 89.2 & \underline{88.3} & 76.8 \\
BO-SA & \underline{80.9} & \underline{89.3} & 87.7 & 66.6 \\
QSA & 64.5 & 72.9 & 88.1 & 76.1 \\
I-QSA & 59.3 & 77.6 & 84.6 & 81.8 \\
\ac{model} (ours) & \textbf{82.5} & \textbf{90.3} &\textbf{92.9} &\textbf{89.2} \\
\bottomrule
\end{tabular}
}
\end{minipage}
\hfill
\begin{minipage}{0.44\linewidth}
\centering
\includegraphics[width=\linewidth]{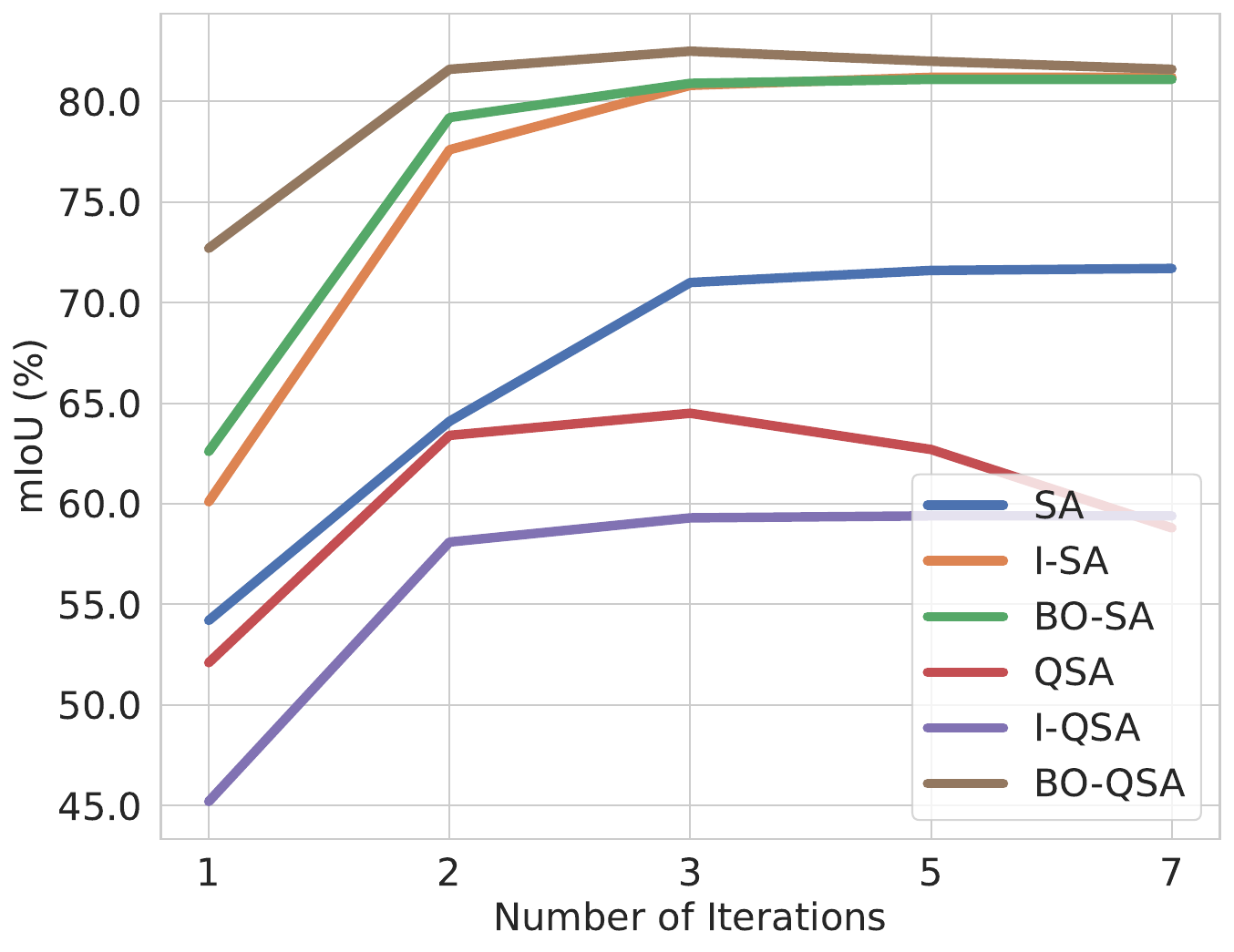}
\captionof{figure}{Effects of iterative updates in testing.}
\label{fig:iteration_inference}
\end{minipage}\\
\begin{minipage}{\linewidth}
\includegraphics[width=\linewidth]{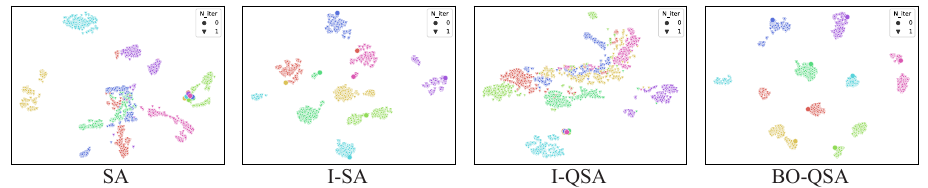}
\captionof{figure}{Visualization of learned slot initializations and post-iteration slots after the first iteration of \sa on ShapeStacks (we use dots for initialization vectors and inverse triangles for post-iteration slots). We show our \ac{model} provides the best overall separation as well as correspondence between initialization vectors and post-iteration slots. For I-SA, there exist mismatches between initialization vectors and post-iteration slots (yellow and red). The same optimization method is also not effective for I-QSA, leading to mixing post-iteration slots similar to SA for slot initializations (best viewed in color and with zoom-in).}
\label{fig:tsne}
\end{minipage}
\end{table}
\paragraph{Experimental Setup} We perform ablative studies over our designs by comparing them with different design variants on ShapeStacks and Stanford Dogs. For slot initialization, we consider (1) the original \sa module's sampling initialization (SA), and (2) initializing with learnable queries (QSA). For optimization, we consider (1) the original optimization in \sa (\ie w/o detach or \ac{ste}), (2) the \ac{isa} optimization where gradients to slots in iterative updates are detached (\ie w/ detach only), and (3) our optimization where we both detach the gradients into iterative refinement, and pass gradient to the initialization queries with \ac{ste} (\ie w/ detach and \ac{ste}). For simplicity, we term these variants with prefixes (I-) for I-SA and (\shortprefix) for our full method. We run all ablations on each dataset with the same encoder-decoder architecture.

\paragraph{Results} We show experimental results in~\cref{tab:ablation} and~\cref{fig:iteration_inference}. First, from~\cref{tab:ablation}, we observe that \ac{model} significantly outperforms other variants. For sample-based slot initializations, our method shows a similar effect compared with~\ac{isa} on improving \sa learning. For query-based slot initializations, we validate the difficulty in training query-based \sa with its inferior performance. We further show the ineffectiveness of~\ac{isa} for query-based \sa. The experiments on query-based \sa prove that both of our design choices are necessary and effective for superior performance. To study the effect of learned queries, we visualize in~\cref{fig:iteration_inference} where we set different numbers of iterative updates of \sa during inference on the Stanford Dogs dataset. We can see that our~\ac{model} significantly outperforms other variants with only one iteration. This indicates that our query-based design can help ease training difficulties. In \cref{fig:tsne}, we further visualize the learned initializations and post-iteration slots in the same feature space using t-SNE~\citep{van2008visualizing}. Our initializers provide a more separable space when differentiating image features, which validates the desired model behaviors mentioned in~\cref{sec:model:boqsa}.

\subsection{Additional Analyses}\label{sec:exp:transfer}
\begin{figure}[t!]
    \centering
   \resizebox{\linewidth}{!}{\includegraphics[width=\linewidth]{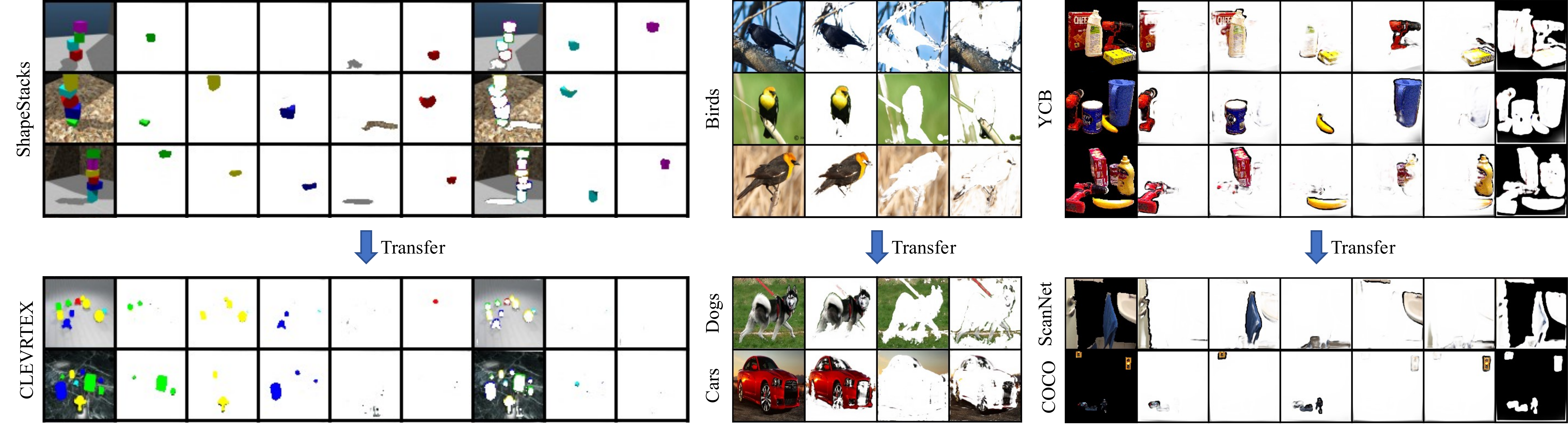}}
    \caption{Visualization of learned concepts and attention maps in zero-shot transfer. At the top, we visualize the per-slot reconstruction of our model trained on ShapeStacks (left), Birds (middle), and YCB (right). At the bottom, we show that our learned slot initialization queries bind to the same concepts in zero-shot transfer experiments  (\ie color in ShapeStacks to CLEVRTEX, contours in Birds to Dogs and Cars, and spatial positions in YCB to ScanNet and COCO) by visualizing attention maps of slot initialization queries over input images. *Note that for the ShapeStacks experiment(left), we alternate object colors in CLEVRTEX with seen colors for better qualitative evaluations, and we do not perform such operations for quantitative evaluations.}
    \label{fig:concept_binding}
\end{figure}
In this section, we provide additional analyses on the potential of our~\ac{model} as a concept binder for generalizing to new examples. First, we qualitatively visualize our learned content for each slot (without additional clustering) in ShapeStacks, Birds, and YCB in~\cref{fig:concept_binding}.
We observe high similarity within the learned content of each slot, indicating similar concepts learned by specific slots. This shows the potential of the slots in our~\ac{model} for binding specific concepts on object properties (\eg colors, contours, and spatial positions). Although we can not control which concepts to learn, these results are important indicators that our learned initialization queries could potentially be generalizable concept probes.
We 
further 
\begin{wrapfigure}[7]{c}{0.5\linewidth}
\raisebox{0pt}[\dimexpr\height-1\baselineskip\relax]{\begin{minipage}{\linewidth}
\captionof{table}{Zero-shot transfer results of unsupervised multi-object segmentation on real images.}
\label{tab:transfer}
\resizebox{\linewidth}{!}{
\begin{tabular}{ccc}
\toprule
\multirow{2}[2]{*}{Model} & YCB $\to$ ScanNet & YCB $\to$ COCO \\
    & (AP / PQ / Pre / Rec) & (AP / PQ / Pre / Rec) \\
    \midrule
    SA & 1.37 / 4.90 / 11.27 / 6.35 & 1.20 / 4.97 / 10.48 / 6.73 \\
    I-SA & 21.62 / 21.81 / 32.32/ 34.19 & 18.39 / 18.47 / 27.23 / 30.38 \\
    \ac{model} (ours) &  \textbf{28.24 / 25.93 / 36.68 / 39.62} &  \textbf{24.23 / 21.65 / 30.20 / 35.79} \\
\bottomrule
\end{tabular}
}
\end{minipage}
}
\end{wrapfigure}
provide quantitative evaluations where we use models trained on dataset X for zero-shot inference on dataset Y.
We term this transfer as (X$\to$Y). As shown in~\cref{tab:transfer}, when adapting models trained on YCB to zero-shot inference on ScanNet and COCO, our method outperform~\ac{isa} and also the majority of fine-tuned methods shown in~\cref{tab:real_multi_seg}. Due to the page limit, we show in~\cref{app:exp:results} that this superior transfer capability is general across datasets when compared to \sa variants.

\section{Conclusions}
We introduce~\ac{model} for unsupervised object-centric representation learning. We initialize \sa with learnable queries, and combine bi-level optimization and straight-through gradient estimators to ease the difficulty in query-based \sa learning. With simple code adjustments on \sa, we obtain \sota model for unsupervised object segmentation in both synthetic and natural image domains, outperforming previous baselines by a large margin. More importantly, our learned model exhibits concept-binding effects where visual concepts are attached to specific slot queries. 
With a fixed number of initialized slots, our model is limited to handling a fixed maximum number of objects in the inputs. However, our queries could be learned to bind object attributes, which leads to meaningful segmentation of images by grouping similar properties (\eg color, position, \etc). As a future direction, this connects our method with weakly-supervised contrastive learning methods that learn grounded visual representations with language.
\newpage
\section*{Acknowledgement}
We gratefully thank all colleagues from BIGAI for fruitful discussions. We would also like to thank the anonymous reviewers for their constructive feedback. This work reported herein was supported by National Key R\&D Program of China (2021ZD0150200).

\bibliography{ref}
\bibliographystyle{iclr2023_conference}

\clearpage
\appendix
\section{Model Architecture and Design}\label{app:model}
\subsection{Design of Decoders}\label{app:model:decoder}
In this section, we follow the notations used in~\cref{sec:prelim:slot_attention} and describe two common approaches, mixture-based and transformer-based, for decoding images from the learned slot representations.

\paragraph{Mixture-based Decoder} The mixture-based decoder~\citep{watters2019spatial} decodes each slot $\hat{\rvs}_i$ into an object image $\rvx_{i}$ and mask $\rvm_i$ with decoding functions $g^{\text{img}}_{\phi^{\text{dec}}}$ and $g^{\text{mask}}_{\phi^{\text{dec}}}$, which are implemented using CNNs. The decoded images and masks are calculated by:
\begin{equation*}
    \begin{aligned}
        \hat{\mI}_i = g_{\phi^{\text{dec}}}^{\text{img}}(\hat{\rvs}_i),&& \rvm_i = \frac{\exp\ g_{\phi^{\text{dec}}}^\text{mask} (\hat{\rvs}_i)}{\sum_{j=1}^{K}\exp\ g_{\phi^{\text{dec}}}^{\text{mask}}(\hat{\rvs}_j)},&& \hat{\mI} = \sum_{i=1}^{K}\rvm_{i}\cdot\hat{\mI}_i.
    \end{aligned}
\end{equation*}
During training, a reconstruction objective is employed for supervising model learning. Despite its wide usage, mixture-based decoders showed limited capability at handling natural scenes with high visual complexity~\citep{singh2021slate}.
\paragraph{Autoregressive Transformer Decoder} Recently,~\cite{singh2021slate,singh2022simple} reveal the limitations of mixture decoder and leverage transformers and \ac{dvae}s~\citep{van2017neural, ramesh2021zero} for decoding slot-based object-centric representations. To obtain decoded images $\hat{\mI}$, they learn a separate \ac{dvae} for first encoding $\mI$ into a sequence of $L$ tokens $\rvz=\{\rvz_1,\cdots,\rvz_L\}$ with \ac{dvae} encoder $f_{\phi^{\text{enc}}}^{\text{dVAE}}$. Next, they use a transformer decoder $g_{\phi^{\text{dec}}}^{\text{transformer}}$ to auto-regressively predict image tokens with learned slot representation $\hat{\rvs}$:
\begin{equation*}
    \begin{aligned}
        \rvo_l = g_{\phi^{\text{dec}}}^{\text{transformer}}(\hat{\rvs}; \rvz_{<l}) && \text{where} && \rvz = f_{\phi^{\text{enc}}}^{\text{dVAE}}(\mI).
    \end{aligned}
\end{equation*}
To train the entire model, we have the reconstruction objective supervising the learning of $\rvz$ with \ac{dvae} decoder $g_{\phi^{\text{dec}}}^{\text{dVAE}}$. Next, the objective for object-centric learning relies on the correct prediction from the auto-regressive transformer for predicting correct tokens:
\begin{equation*}
    \begin{aligned}
        \mathcal{L} = \mathcal{L}_{\text{dVAE}} + \mathcal{L}_{\text{CE}} && \text{where} && \mathcal{L}_{\text{dVAE}} = ||g_{\phi^{\text{dec}}}^{\text{dVAE}}(\rvz) - \mI||_2^2,\ \mathcal{L}_{\text{CE}} = \sum_{l=1}^{L}\text{CrossEntropy}(\rvz_l, \rvo_l)
    \end{aligned}
\end{equation*}
Under this setting, the model does not predict additional masks and relies on the attention $\mA$ within the \sa module for obtaining slot-specific object masks. Although such models can achieve competitive results on real-world synthetic datasets, as our experiments suggest, they can be inferior to mixture-based decoders on segmentation in synthetic datasets. We suspect that this originates from the low resolution when discretizing images into tokens.

\subsection{Bi-level Optimization and Meta-Learning}\label{app:model:optimization}
Recall the bi-level optimization problem we introduced in~\cref{sec:prelim:meta_learning}.

\begin{equation}
    \begin{aligned}
    \min_{\theta, \phi} f(\theta, \phi) && s.t. && \theta \in \argmin_{\theta'}g(\theta', \phi),
    \end{aligned}
    \label{eq:bo}
\end{equation}

where we call $f(\theta,\phi)$ the outer objective function and $g(\theta, \phi)$ the inner objective function. To jointly optimize both objectives \wrt parameters $\theta$ and $\phi$, a straightforward approach to solving~\cref{eq:bo} is to represent the inner solution of $\theta$ as a function of $\phi$, \ie, $\theta^*(\phi) = \argmin_{\theta'}g(\theta', \phi)$. Then we can optimize the outer objective with gradient descent:
\begin{equation*}
    \nabla_{\phi}f(\theta^*(\phi), \phi) = \nabla_{\phi}\theta^*(\phi)\nabla_1f(\theta^*(\phi), \phi) + \nabla_2 f(\theta^*(\phi), \phi),
\end{equation*}

However, the difficulty of this method lies in the calculation of $\nabla_{\phi}\theta^*(\phi)$ where we need to solve linear equation from implicit gradient theorem:
\begin{equation*}
    \nabla_{1,2}g(\theta^*(\phi),\phi)\nabla_\phi\theta^*(\phi) + \nabla_{2,2}g(\theta^*(\phi), \phi) = 0.
\end{equation*}
If $\nabla_{2,2}g(\theta^{*},\phi)$ is invertible, we can solve for $\nabla_{\phi}\theta^{*}(\phi)$ and obtain the gradient update on $\phi$:
\begin{equation*}
    \phi_{k+1} = \phi_{k} - \xi\left(\nabla_2f_{k} - (\nabla_{1,2}g_k)^{\top}(\nabla_{2,2}g_k)^{-1}\nabla_{1}f_k\right)
\end{equation*}
where $\nabla_1 f_k =\nabla_2f(\theta^{*}(\phi_k),\phi_k)$ and $\nabla_1f_k = \nabla_{1}f(\theta^{*}(\phi_{k}), \phi_k)$. Various methods have been proposed to approximate the solution~\citep{pedregosa2016hyperparameter, lorraine2020optimizing}, and we refer the authors to~\cite{ye2022bome} for a thorough review of related methods.

Bi-level optimization is closely related to \meta. In \meta, we have meta-training tasks which comes in as $N$ different collections of datasets $\mathcal{D} = \{\mathcal{D}_i = \mathcal{D}_i^{\text{tr}} \cup \mathcal{D}_i^{\text{val}}\}_{i=1}^{N}$. The inner and outer objectives in~\cref{eq:bo} are substituted by averaging training and validation errors over multiple tasks~\citep{franceschi2018bilevel}:
\begin{equation}
    \begin{aligned}
     \min_{\theta,\phi} f(\theta, \phi) = \sum_{i=1}^{N}\mathcal{L}_i(\theta_i, \phi, \mathcal{D}^{\text{val}}_{i}) && s.t. && \theta_i = \min_{\theta_i'} \sum_{i=1}^{N}\mathcal{L}_i(\theta_i', \phi; \mathcal{D}^{\text{tr}}_i), &&
    \end{aligned}
    \label{eq:meta_learning}
\end{equation}
where $\mathcal{L}_i$ represents task-dependent error on $\mathcal{D}_i$. The final goal of \meta aims at seeking the meta-parameter $\phi$ that is shared between tasks which later enables few-shot learning and fast adaptation. With its connections with bi-level optimization, the previously mentioned optimization methods are broadly adapted for solving \meta problems~\citep{finn2017model, nichol2018reptile,rajeswaran2019meta}. From the \meta perspective, our attempt shares similar insights with first-order \meta methods~\citep{finn2017model,nichol2018reptile}, where we use the gradient at some task-specific optimal solution $\rvs_{i}^*$ of the inner optimization for optimizing slot initialization queries which are shared across datasets on the outer objective. This \meta perspective also indicates the potentials of our~\ac{model} for fast adaptation and generalization. 

\subsection{Implementation Details}\label{app:model:implement}
\begin{figure}[t!]
    \centering
     \includegraphics[width=\linewidth]{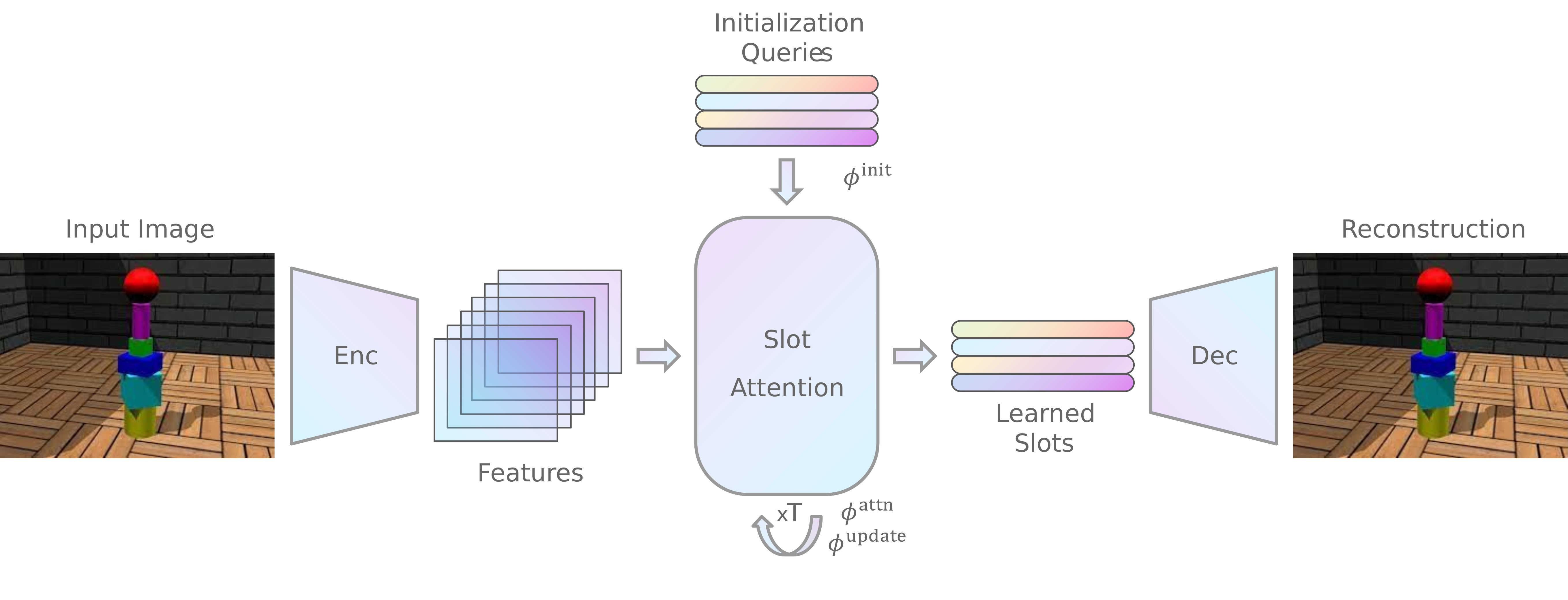}
    \caption{An illustrative visualization of our proposed~\ac{model} slot-encoder. During the backward pass,~\ac{model} uses \ac{ste} to backpropagate gradients directly to $\phi^{\text{init}}$, $\phi^{\text{attn}}$, and $\phi^{\text{update}}$ without gradients into the iterative process.}
    \label{fig:overview}
\end{figure}

We provide a visualization of our designed slot-encoder in~\cref{fig:overview} and discuss the implementation details for different experimental settings in the following sections.

\subsubsection{Slot Initialization} We initialize all models with the number of slots shown in~\cref{tab:num_slots}. During training, we add a small perturbation to the
queries by sampling from a zero-mean distribution with variance $\sigma$ as we found it empirically helpful for better performance. We perform annealing over $\sigma$ to gradually eliminate the effect of this random perturbation during training. We adopt the cosine annealing strategy such that $\sigma$ starts from 1 and gradually anneals to 0 after $N_\sigma$ training steps, where $N_\sigma$ is a hyperparameter that controls the annealing rate of $\sigma$. In our experiments, we use $N_\sigma=0$ on Cars and Flowers and $N_\sigma=30000$ on the rest of the datasets.

\subsubsection{BO-QSA with Mixture-Based Decoders}
For mixture-based decoders, we use the same \sa architecture as in~\cite{locatello2020sa} with slots initialized by learnable queries. Given an input image, \sa uses a CNN encoder to extract image features. After adding positional embedding, these features are input into the \sa module slot updates. Finally, these slots are decoded by the mixture decoder to reconstruct the input image. We provide the details of our image encoder in~\cref{tab:CNN encoder}. For the mixture-based decoder, we use six transposed convolutional layers with ReLU activations following~\cite{locatello2020sa}. We visualize the details of our mixture-based decoder design in~\cref{tab:CNN decoder}. We train our model for 250k steps with a batch size of 128 and describe all training configurations and hyperparameter selection~\cref{tab:training mix}.
\begin{table*}[t]
\centering
\begin{tabular}{cccccc}
\toprule
Layer &Kernel Size &Stride &Padding &Channels &Activation \\
\midrule
Conv &5x5 &1(2) &2 &64 &ReLU \\
Conv &5x5 &1 &2 &64 &ReLU \\
Conv &5x5 &1 &2 &64 &ReLU \\
Conv &5x5 &1 &2 &64 &ReLU \\
\bottomrule
\end{tabular}
\caption{Configuration of CNN encoder used in our model. The values in parentheses are adopted for CLEVRTex and ShapeStacks} \label{tab:CNN encoder}
\end{table*}

\begin{table*}[t]
\centering
\begin{tabular}{cccccc}
\toprule
Layer &Kernel Size &Stride &Padding &Channels &Activation \\
\midrule
TransConv &5x5 &2 &2 &64 &ReLU \\
TransConv &5x5 &2 &2 &64 &ReLU \\
TransConv &5x5 &2 &2 &64 &ReLU \\
TransConv &5x5 &2(1) &2 &64 &ReLU \\
TransConv &5x5 &1 &2 &64 &ReLU \\
TransConv &3x3 &1 &1 &4 &None \\
\bottomrule
\end{tabular}
\caption{Configuration of mixture decoder used in our model. The values in parentheses are adopted for ObjectsRoom} \label{tab:CNN decoder}
\end{table*}

\begin{table*}[ht!]
\centering
\begin{tabular}{cccc}
\toprule
Batch Size &LR &Slot Dim &MLP Hidden Dim\\
128 &4e-4 &64 &128\\
\midrule
Warmup Steps &Decay Steps &Max Steps &Sigma Down Steps\\
5k &50k &250k &30k \\
\bottomrule
\end{tabular}
\caption{Training configuration for mixture-based model} \label{tab:training mix}
\end{table*}

\subsubsection{BO-QSA with Transformer-Based Decoder}
For transformer-based decoders, we adopt the transformer architecture proposed by SLATE~\citep{singh2021slate}. For the transformer-based~\ac{model}, unlike SLATE, we use the same CNN as in mixture-based \ac{model} (instead of the \ac{dvae} encoder) to extract features from the image as input to the \sa module as we find such changes help solve the problem on coarse object boundary prediction mentioned in~\cref{sec:exp:synthetic}. Next, we use the same overall architecture of \ac{dvae} as mentioned in SLATE ~\cite{singh2021slate}. However, we change the kernel size of the \ac{dvae} encoder from 1 to 3 since we find that such changes can help increase model performance when decomposing scenes. We train our model for 250k steps with a batch size of 128, and all the training configuration in our experiments is described in~\cref{tab:trans_config}.

\begin{table*}[ht!]
\centering
\begin{tabular}{ccc}
\toprule
\multirow{4}{*}{Training} & batch size &128 \\
& warmup steps  &10000 \\
& learning rate  &1e-4 \\
& max steps  &250k \\
\midrule
\multirow{4}{*}{\ac{dvae}} &vocabulary size &1024 \\
&Gumbel-Softmax annealing range &1.0 to 0.1 \\
&Gumbel-Softmax annealing steps &30000 \\
&lr-\ac{dvae}(no warmup)  &3e-4 \\
\midrule
\multirow{4}{*}{Transformer Decoder} &layers &4 \\
&heads &4 \\
&dropout &0.1 \\
&hidden dimension &256 \\
\midrule
\multirow{3}{*}{Slot Attention Module} & slot dimension &256 \\
&iterations &3 \\
& $\sigma$ annealing steps &30000(0) \\
\bottomrule
\end{tabular}
\caption{Training configuration for transformer-based model. The values in parentheses are adopted for Cars and Flowers dataset} \label{tab:trans_config}
\end{table*}

\subsubsection{Baselines}
The reproduction of \sa and SLATE follows the architecture and hyperparameter selection mentioned in their paper. Similar to our models, we train all baseline models with 250K steps on all datasets. For SLATE, we use the input image size of 96 on the ShapeStacks dataset as we find that the image size of 128 will cause all objects to be divided into the same slot, resulting in low ARI and MSC. For a fair comparison with numbers reported in SLATE's paper, we report the MSE of models by first computing per-pixel errors and then multiplying it by the total number of pixels. For CLEVRTEX, we follow the same experimental setting of (\ac{model}+mixture) for ShapeStacks and set the number of slots to 11. For YCB, ScanNet, and COCO, we follow the same experimental setting of (\ac{model}+transformer) for birds and set the number of slots to 6.
\begin{table*}[ht!]
\centering
\resizebox{\linewidth}{!}{
\begin{tabular}{ccccccccc}
\toprule
&Model &Shapestacks &ObjectsRoom &Birds &Dogs &Flowers &Cars \\
\midrule
\multirow{4}{*}{\# of slots}
&\sa &8 &5 &3 &2 &2 &2 \\
&SLATE &12 &6 &3 &2 &2 &2 \\
&\abbrev+Mixture &8 &5 &3 &2 &2 &2 \\
&\abbrev+Transformer &12 &6 &3 &2 &2 &2 \\
\midrule
& Image Size &128 &64 &128 &128 &128 &128 \\
\bottomrule
\end{tabular}
}
\caption{The number of slots and image size used for each dataset} \label{tab:num_slots}
\end{table*}

\section{Additional Experiments}
\subsection{Zero-shot Transfer}\label{app:exp:results}
In this section, we continue the discussion in~\cref{sec:exp:transfer} and provide additional zero-shot transfer results. Similarly, we use the notation ($X\rightarrow Y$) to denote the zero-shot adaptation of models trained unsupervisedly on dataset $X$ to new datasets $Y$. 

For unsupervised multi-object segmentation, we report transfer results from ScanNet and COCO to all other real-image multi-object segmentation datasets in addition to the results on YCB (mentioned in~\cref{sec:exp:transfer}). As shown in~\cref{tab:add_transfer_multi}, our model shows consistent improvement over \sa and \ac{isa} during zero-shot transfer.

\begin{table}[ht!]
\caption{Zero-shot transfer results of unsupervised multi-object segmentation on real images.}
\label{tab:add_transfer_multi}
\resizebox{\linewidth}{!}{
\begin{tabular}{ccccc}
\toprule
\multirow{2}{*}{Model} & ScanNet $\to$ YCB & ScanNet $\to$ COCO & COCO $\to$ YCB 
& COCO $\to$ ScanNet \\
& (AP / PQ / Pre / Rec) $\uparrow$ & (AP / PQ / Pre / Rec) $\uparrow$ & (AP / PQ / Pre / Rec) $\uparrow$ & (AP / PQ / Pre / Rec) $\uparrow$\\ 
    \midrule
    SA & 19.63 / 19.24 / 28.56 / 31.43 & 12.84 / 14.86 / 22.06 / 26.74 & 26.53 / 23.05 / 35.96 / 38.12 & 20.99 / 22.08 / 32.14 / 36.53 \\
    I-SA & 18.66 / 18.56 / 28.97 / 30.82 & 11.83 / 14.14 / 20.70 / 25.42 & 26.72 / 22.90 / 35.89 / 37.98 & 19.34 / 20.00 / 29.44 / 33.18\\
    \abbrev (ours) & \textbf{21.85 / 19.96 / 31.51 / 33.45} & \textbf{13.95 / 16.04 / 23.35 / 28.49} & \textbf{31.21 / 25.44 / 38.90 / 41.35} & \textbf{24.21 / 23.59 / 34.07 / 38.49}\\
\bottomrule
\end{tabular}
}
\end{table}

For unsupervised foreground extraction, we report transfer results from Stanford Dogs and CUB200 Birds to all other real-image foreground extraction datasets. As we can see from~\cref{tab:add_transfer}, our model achieves the overall best results compared with other powerful \sa variants (models that achieve best or second-best results in our ablation studies as in~\cref{tab:ablation}) except for (Birds$\rightarrow$Cars). However, our optimization method still helps improve zero-shot transfer for randomly initialized \sa.
\begin{table}[ht!]
\caption{Zero-shot transfer results on unsupervised foreground extraction (mIoU $\uparrow$).}
\label{tab:add_transfer}
\resizebox{\linewidth}{!}{
\begin{tabular}{ccccccc}
\toprule
Model & Dogs $\to$ Cars & Dogs $\to$ Flowers & Dogs $\to$ Birds 
& Birds $\to$ Dogs & Birds $\to$ Cars & Birds $\to$ Flowers \\
    \midrule
    SA &57.96 &57.96 &45.06 &74.68 &58.79 &62.02 \\
    I-SA &58.05 &58.06 &48.88 &71.16 &69.90 &68.67 \\
    BO-SA &58.10 &58.10 &47.96 &71.81 &\textbf{70.75} &67.95 \\
    \abbrev (ours) &  \textbf{75.50} &  \textbf{63.43} &  \textbf{52.49}
    &  \textbf{76.66} &  66.74 &  \textbf{70.74} \\
\bottomrule
\end{tabular}
}
\end{table}

\subsection{Analysis Number of \sa Iterations}\label{app:exp:iqsa}
As described in~\cref{sec:model:meta_learning}, we study whether a fixed point $\rvs^*$ could be reached by a fixed number of iterations during training. Since we hypothesized that the low performance of I-QSA in~\cref{sec:exp:ablation} originated from the insufficient number of starting points for fixed-point approximation, we conduct experiments on increasing the number of \sa iterations during training for I-QSA on the Dog dataset. As shown in~\cref{tab:add_iteration}, increasing the number of \sa iterations during training for I-QSA significantly improves its performance. However, we found that adding more iterations after a threshold (\ie 7 in this case) does not further improve the overall performance. This verifies the need for learning slot initialization vectors for better approximating the fixed point solution of the inner soft-clustering objective in \sa.
\begin{table}[ht!]
\caption{Increasing the number of iterations during training for I-QSA.}
\label{tab:add_iteration}
\centering
\begin{tabular}{cccccc}
\toprule
\multirow{2}[2]{*}{Model} & \multirow{2}[2]{*}{\# of Training Iterations} & \multicolumn{4}{c}{Dogs} \\
\cmidrule(lr){3-6}
& & $\uparrow$ IoU & Gain & $\uparrow$ Dice & Gain\\
\midrule
I-QSA & 3 & 59.3 & - & 77.6 & -\\
I-QSA & 7 & 80.5 & +35.8\%& 88.9 & +14.6\%\\
\midrule
Ours & 3 & \textbf{82.5} & - & \textbf{90.3} & -\\
\bottomrule
\end{tabular}
\end{table}

\subsection{Design Choices on Slot Initialization}\label{app:exp:dvae}
As described in~\cref{sec:model:boqsa}, our method is connected with recent works on~\ac{dvae}. However, we do not require the initialization queries to maintain information about the post-iteration slots $\hat{\rvs}$ as we found such constraints lead to the learning of the mean representation of datasets which forbids disentanglement and concept binding. In this section, we provide experimental results to verify this argument. Specifically, we consider three different ways to update slot initialization queries in addition to our proposed method: 1) using the running mean of the post-iteration slots as initialization queries (RunningMean), 2) running K-Means clustering on post-iteration slots and updating the initialization queries using re-clustered centers by Hungarian matching (KMeans), 3) adding consistency loss between initialization queries and post-iteration slots as done in VQ-VAE (VQ-constraint). For (1) and (2), we empirically found such designs to be suffering from frequent updates and therefore use momentum updates to stabilize their training. We term these variants with the suffix (-M).

\begin{table}[ht!]
\caption{Comparison between update methods for slot-initialization queries.}
\label{tab:add_initialization}
\centering
\resizebox{\linewidth}{!}{
\begin{tabular}{ccccccc}
\toprule
Metrics &  RunningMean & RunningMean-M & KMeans & KMeans-M & VQ-constraint & Ours\\
\midrule
ARI-FG (ShapeStacks) &7.5 &51.4 &21.0 &70.6 &88.6 &\textbf{92.9}\\
MSC-FG (ShapeStacks) &3.7 &15.4 &4.2 &60.4 &85.3 &\textbf{89.2}\\
\bottomrule
\end{tabular}
}
\end{table}

\begin{figure}[t!]
    \resizebox{\linewidth}{!}{
        \includegraphics[width=\linewidth]{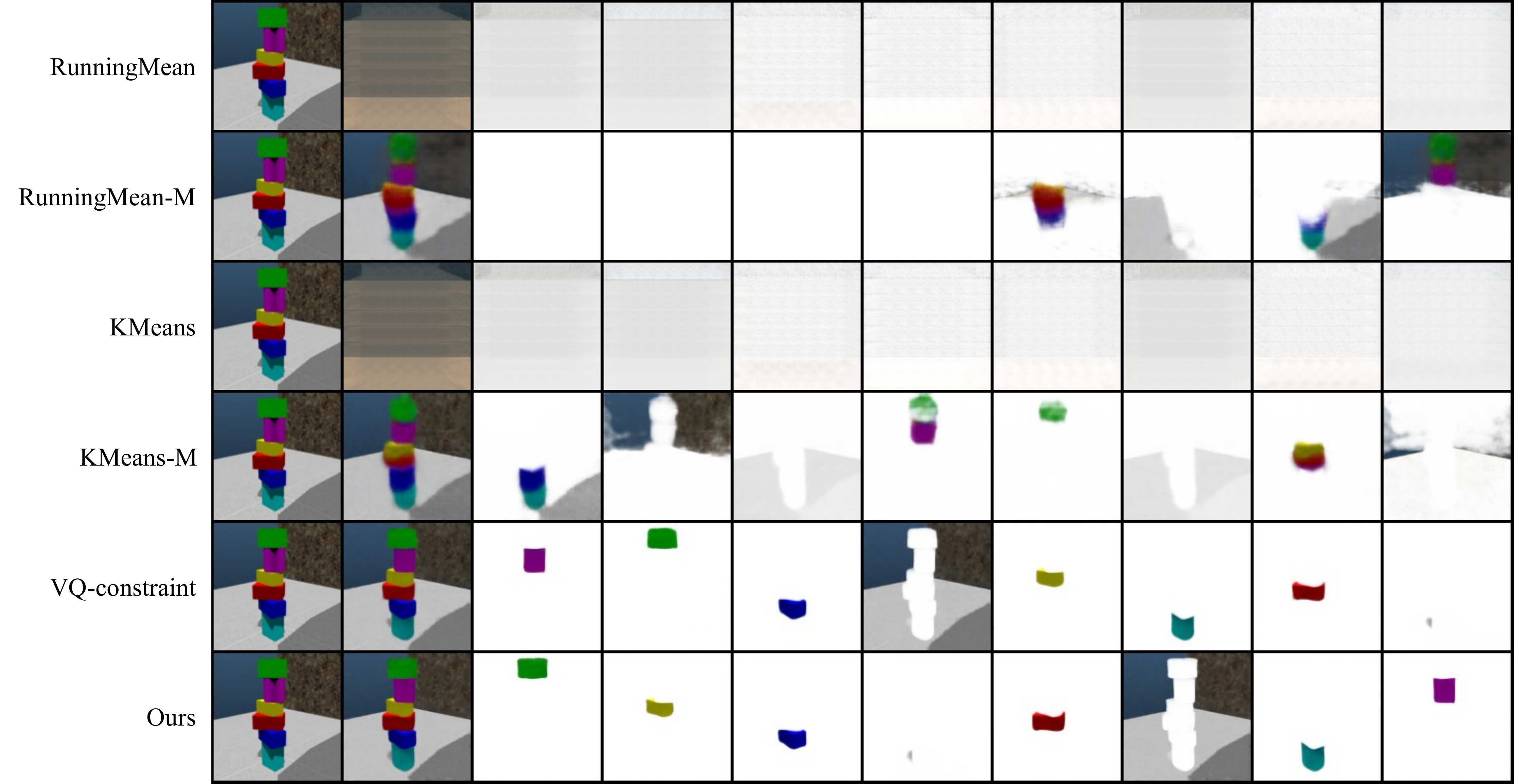}
    }
    \caption{Visualizations per-slot reconstruction for different update methods. We show that RunningMean and KMeans suffer at decomposing the image, even with momentum updates. For VQ-constraint, though the model variant achieves a similar but slightly inferior effect on segmentation, they can not preserve the same filtered property for each slot across images.}
    \label{fig:add_init_ablation}
\end{figure}

    As shown in~\cref{tab:add_initialization}, our model achieves the best overall performance compared to other initialization methods. Specifically, we found that using the running mean of post-iteration slots or K-Means cluster centers re-clustered from post-iteration slots to be harmful to model performance. We attribute this effect to the learning of the mean-representation of datasets. This is further proved in experiments with VQ-VAE loss on consistency between slot initializations and post-iteration slots (\ie $||\text{sg}(\hat{\rvs}) - \rvs_0||^2$), where the VQ-constraint variant showed inferior performance. We also found that the weight of this additional loss needs to be carefully tuned for the model to decompose objects. Empirically, most configurations of this hyperparameter will lead to bad reconstructions except for certain small weights (\eg 0.01 reported here). Above all, we believe these experimental results verify the effectiveness of our design choices on initialization query learning. We provide additional visualizations on the learned contents of slots for each update method in~\cref{fig:add_init_ablation}.

\subsection{Experiments on Additional Datasets}
In addition to datasets considered in~\cref{sec:exp}, we conduct experiments on other synthetic datasets and visualize qualitative results. More specifically, we test our model on PTR~\citep{hong2021ptr}. PTR is a synthetic dataset of 3D objects from PartNet with rendering variations. We run our~\ac{model} with the same configuration mentioned in~\cref{app:model:implement} previously. We compare our method with the vanilla \sa module on multi-object segmentation. We report ARI-FG and MSC-FG scores of our model compared with the vanilla \sa on the PTR validation set. 
\begin{table}[ht!]
\centering
\caption{Multi-object segmentation results on PTR. We visualize the best results in bold.}
\label{tab:syn_add_ptr}
\begin{tabular}{ccc}
    \toprule
    \multirow{2}[2]{*}{Model} & \multicolumn{2}{c}{PTR} \\
    \cmidrule(lr){2-3}
    & ARI-FG $\uparrow$ & MSC-FG $\uparrow$\\
    \midrule
    \sa & 0.72 & 0.21 \\
    ours (\ac{model}+mixture) & \textbf{0.75} & \textbf{0.61}\\
    \bottomrule
\end{tabular}
\end{table}

As we can see from~\cref{tab:syn_add_ptr}, our model achieves similar performance compared with \sa on ARI-FG and significantly outperforms it on MSC-FG. We attribute this result to the capability of precisely segmenting objects. As ARI-FG applies masks to each slot prediction for calculating results, it does not require models to precisely segment the object from the background. However, MSC-FG uses a mIoU-like measure that requires the model to precisely predict the object boundaries. This indicates that our model is better at precisely segmenting objects without noise. Similarly, we observe the binding of certain slots to scene backgrounds, but with more complex concepts, the binding of slots to concepts is not as straightforward as in ShapeStacks and CUB200 Birds.

To further investigate the effectiveness and generality of our method, we adapt BO-QSA to the recent 3D object-centric learning model, uORF~\citep{yu2021unsupervised}, and test it on 3D datasets including CLEVR-567, Room-Chair, and Room-Diverse. uORF can decompose complex 3D scenes from a single image by combining NeRF~\citep{mildenhall2021nerf} with \sa. We only modify the initialization and optimization method of the \sa module in uORF, leaving all other hyperparameters unchanged. As we can see from~\cref{tab:uorf}, with our method, the uORF model that trained with 600 epochs can achieve a similar or even superior result compared to the original model trained with 1200 epochs. Additionally, when the dataset complexity increases (\eg, in Room-Diverse), our method demonstrates significant improvement. Please refer to uORF~\citep{yu2021unsupervised} for more details about the model, datasets, and evaluation metrics.

\begin{table}[ht!]
\centering
\caption{3D-object segmentation results on CLEVR-567, Room-Chair, and Room-Diverse. We visualize the best results in bold and underline the second-best results. $^*$indicates reimplemented results.}
\label{tab:uorf}
\resizebox{\linewidth}{!}{
\begin{tabular}{ccccccccc}
    \toprule
    Dataset & Model & Train-epoch & NV-ARI$\uparrow$ & ARI$\uparrow$ & ARI-FG$\uparrow$ & LPIPS$\downarrow$ & SSIM$\uparrow$ & PSNR$\uparrow$ \\
    \midrule
    \multirow{4}[2]{*}{CLEVR-567} & \multirow{2}[1]{*}{uORF} &600$^*$ &66.8 &73.8 &81.0 &0.1249 &0.8763 &27.84 \\ 
    & &1200{\ \ } &\textbf{84.4} &\textbf{87.4} &85.3 &0.0869 &0.8985 &29.32 \\ 
    \cmidrule(lr){2-9}
    & \multirow{2}[1]{*}{uORF+BO-QSA} &600{\ \ } &74.5 &82.9 &\underline{89.1} &\underline{0.0783} &\underline{0.9153} &\underline{30.07} \\ 
    & &1200{\ \ } &\underline{77.7} &\underline{86.9} &\textbf{89.5} &\textbf{0.0711} &\textbf{0.9223} &\textbf{30.64} \\ 
    \midrule
    \multirow{4}[2]{*}{Room-Chair} & \multirow{2}[1]{*}{uORF} &600$^*$ &37.9 &39.4 &18.8 &0.2932 &0.7734 &25.08 \\ 
    & &1200{\ \ } &\underline{77.9} &\underline{80.3} &91.8 &0.0845 &0.8762 &29.66 \\ 
    \cmidrule(lr){2-9}
    & \multirow{2}[1]{*}{uORF+BO-QSA} &600{\ \ } &76.9 &79.8 &\textbf{94.6} &\underline{0.0821} &\underline{0.8850} &\underline{30.13} \\ 
    & &1200{\ \ } &\textbf{80.5} &\textbf{83.2} &\underline{93.8} &\textbf{0.0733} &\textbf{0.8938} &\textbf{30.61} \\ 
    \midrule
    \multirow{4}[2]{*}{Room-Diverse} & \multirow{2}[1]{*}{uORF} &120$^*$ &51.2 &60.1 &62.0 &0.2139 &0.6905 &25.21 \\ 
    & &240{\ \ } &56.6 &68.5 &66.7 &0.1820 &0.7146 &25.92 \\ 
    \cmidrule(lr){2-9}
    &\multirow{2}[1]{*}{uORF+BO-QSA} &120{\ \ } &\underline{60.4} &\underline{70.0} &\underline{75.1} &\underline{0.1657} &\underline{0.7137} &\underline{26.38} \\ 
    & & 240{\ \ } &\textbf{63.0} &\textbf{72.8} &\textbf{76.6} &\textbf{0.1533} &\textbf{0.7378} &\textbf{26.85} \\ 
    \bottomrule
\end{tabular}
}
\end{table}

\section{Limitations and Future Work}
We discuss all limitations of our work found in the experiments. First, we observed a strong correlation between the powerfulness of encoder-decoder architectures and model performance. However, in contrast to supervised learning, more powerful encoders/decoders do not guarantee superior performance. Gaining insights from how contrastive learning methods have shown the effect of concept emergence with large-scale pretraining, we can also incorporate such representations learned by self-supervised learning into object-centric learning to unite the best of both worlds. Second, our work is primarily limited by the fixed number of slot initialization vectors. In contrast to the vanilla~\sa that could generalize to a new number of objects, our model can not easily generalize to scenarios with new concepts since our model learns a fixed set of separating spaces that best disentangle different parts of the image. This problem is also frequently met in semantic segmentation and object classification, where we can only use existing concepts to interpret novel objects/semantic entities. Although solutions to this close-vocabulary problem have been proposed in supervised classification and segmentation, we leave the exploration of this problem in object-centric learning to future work. Finally, the current learned slot initialization vectors do not explicitly bind towards concepts and need to be mined by humans. We believe this is an important next step in our current work to combine unsupervised object-centric learning with semantic alignments from language for concept grounding. This opens future research directions on learning finer-level organization of object concepts under more complex scenarios (\eg hierarchical grouping) with weak supervision of correspondence.

\newpage

\section{Additional Visualizations}\label{app:exp:vis}
We provide more qualitative results of our model on different datasets in the following pages.

\begin{figure}[ht!]
    \centering
    \resizebox{\linewidth}{!}{\includegraphics{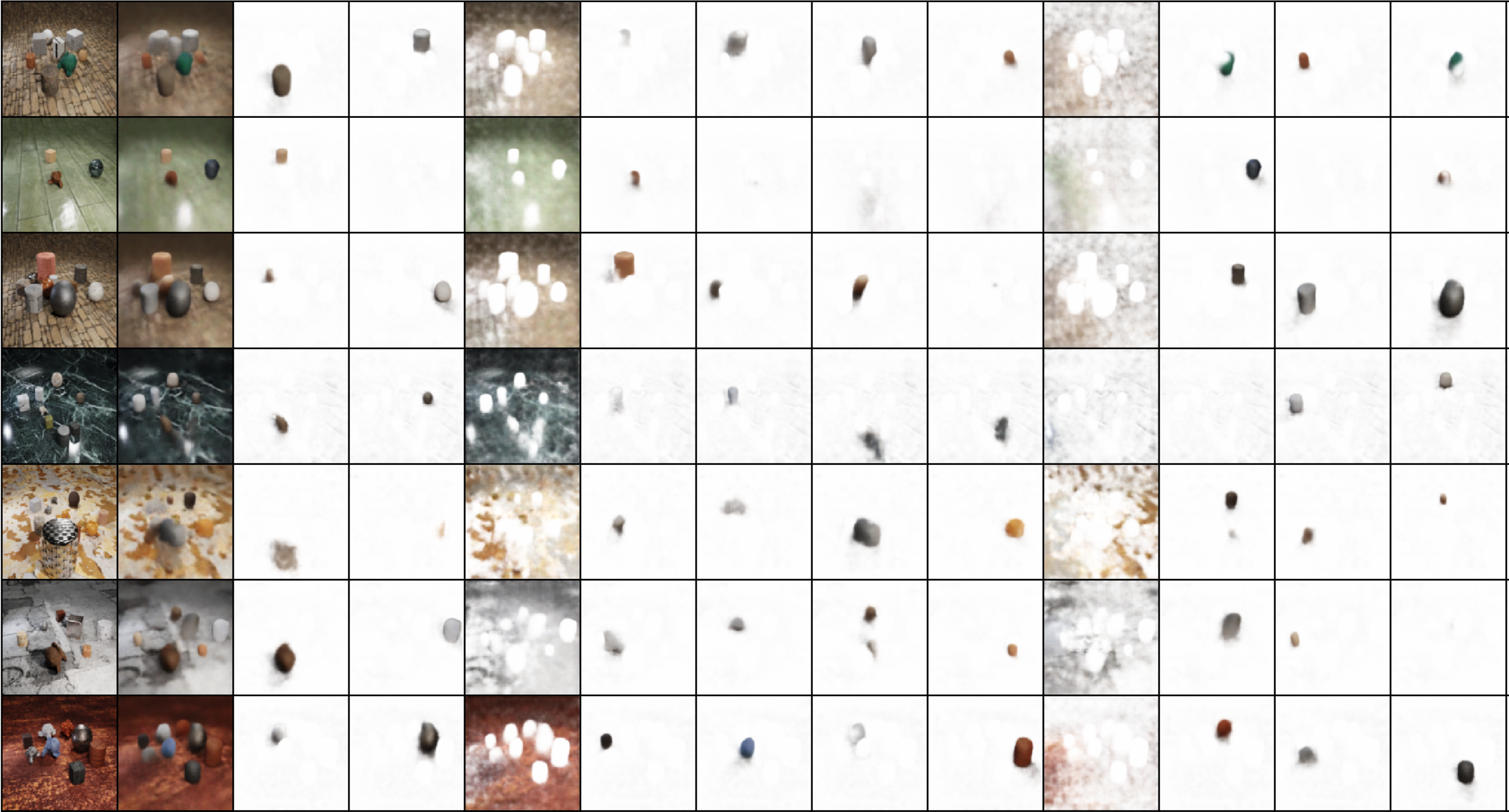}}
    \caption{Unsupervised Multi-Object Segmentation on CLEVRTEX.}
    \label{fig:add:clevrtex}
\end{figure}
\begin{figure}[ht!]
    \centering
    \resizebox{\linewidth}{!}{\includegraphics{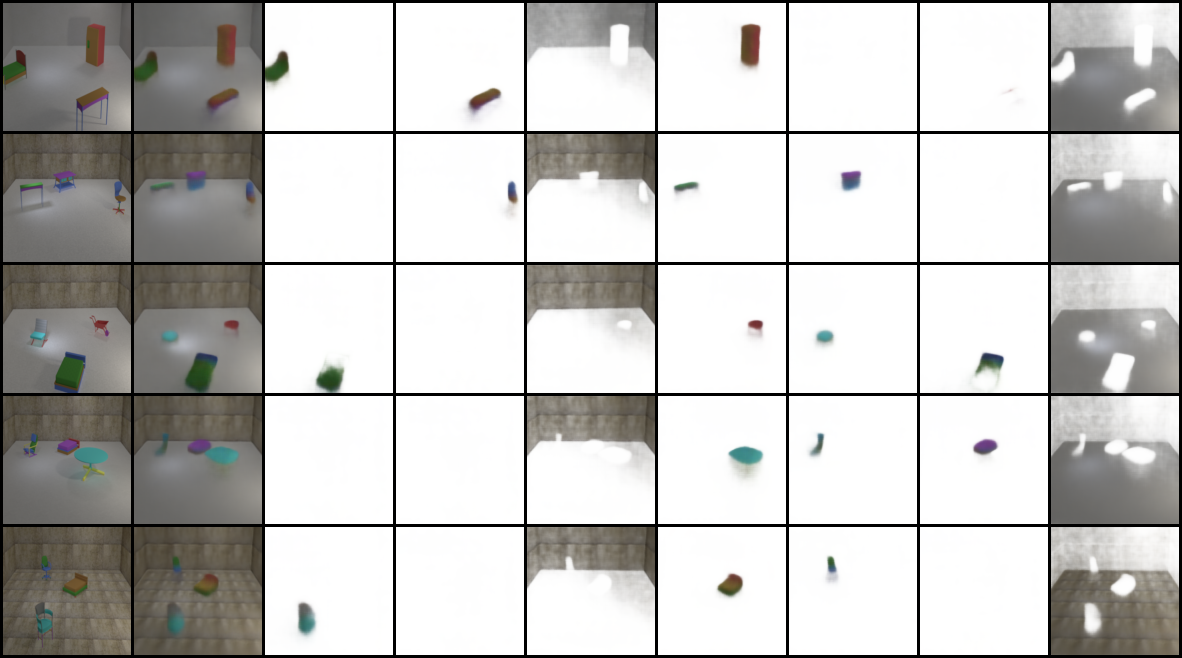}}
    \caption{Unsupervised Multi-Object Segmentation on PTR.}
    \label{fig:add:ptr}
\end{figure}
\begin{figure}[ht!]
    \centering
    \resizebox{0.93\linewidth}{!}{\includegraphics{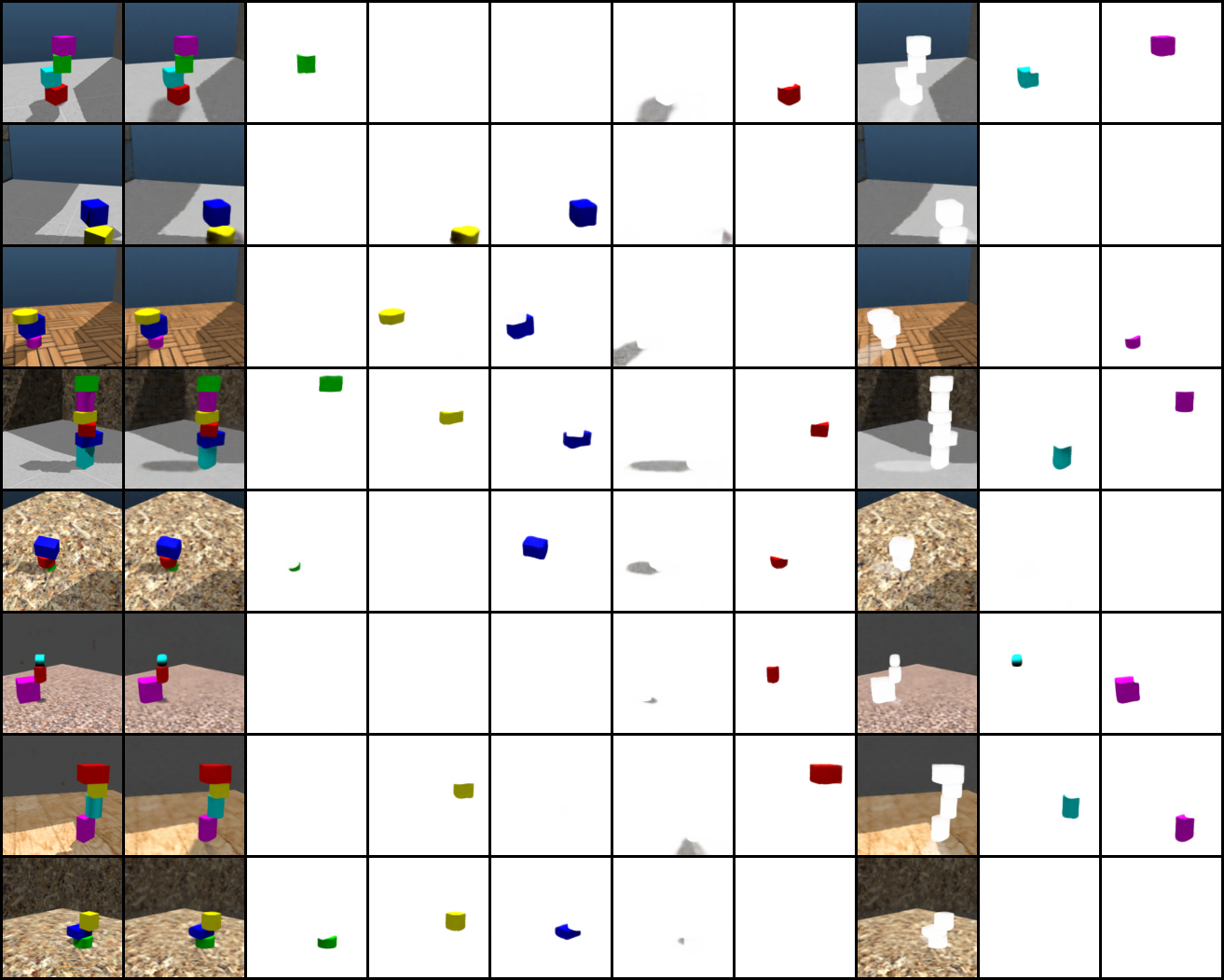}}\\
    \resizebox{0.93\linewidth}{!}{\includegraphics{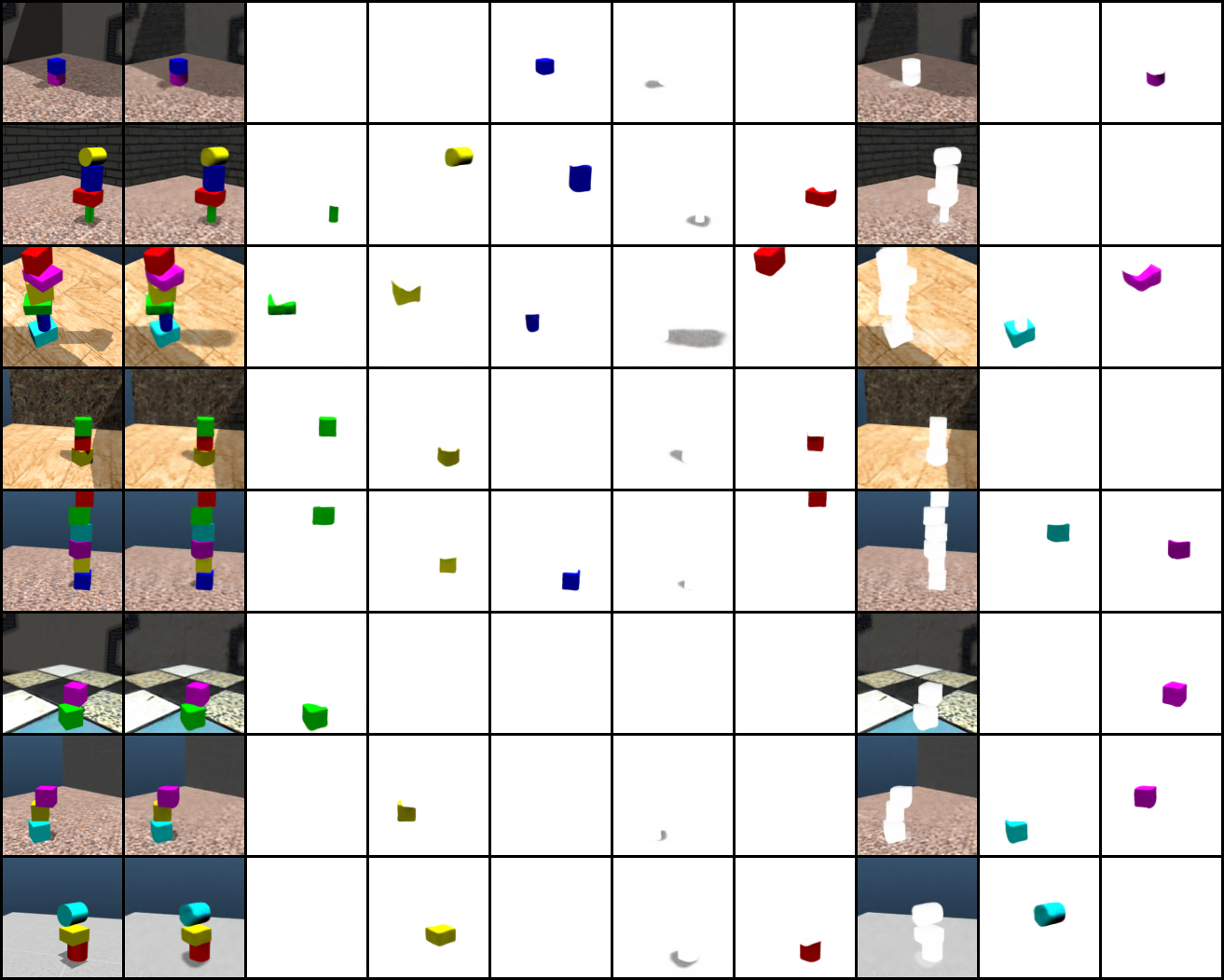}}
    \caption{Unsupervised Multi-Object Segmentation on ShapeStacks.}
    \label{fig:add:shapestacks}
\end{figure}
\begin{figure}[ht!]
    \centering
    \resizebox{0.49\linewidth}{!}{\includegraphics{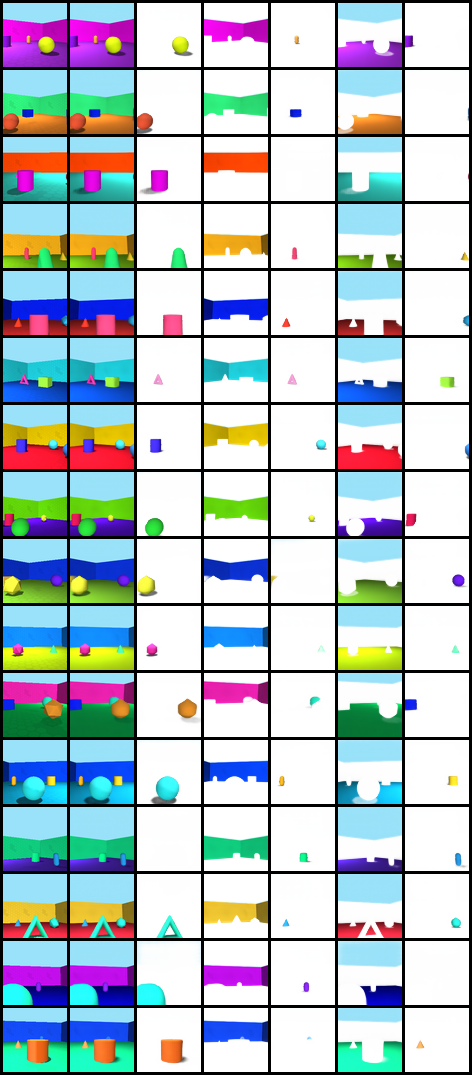}}\hfill\resizebox{0.49\linewidth}{!}{\includegraphics{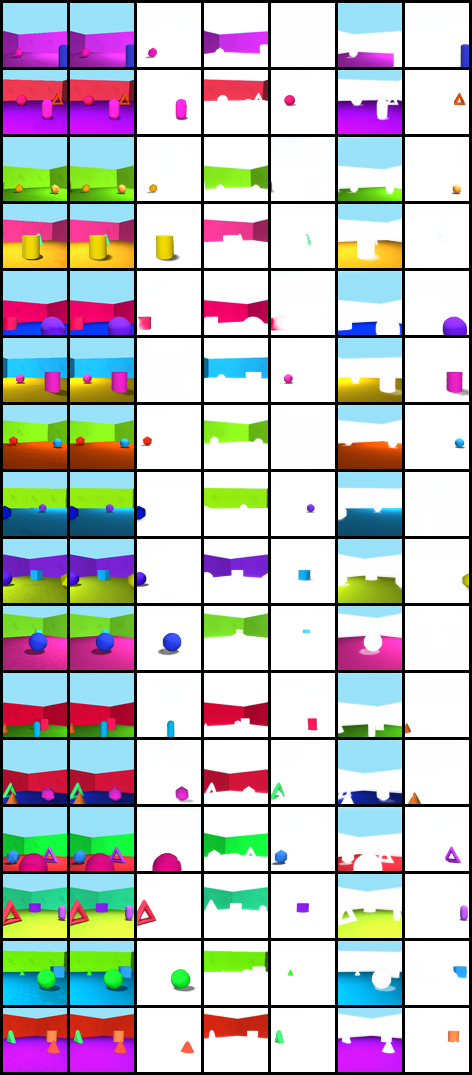}}
    \caption{Unsupervised Multi-Object Segmentation on ObjectsRoom. In contrast to ShapeStacks, we observe consistent binding of slots to ground, wall, sky, and also objects in the front.}
    \label{fig:add:objectsroom}
\end{figure}

\begin{figure}[ht!]
    \centering
    \resizebox{0.5\linewidth}{!}{\includegraphics{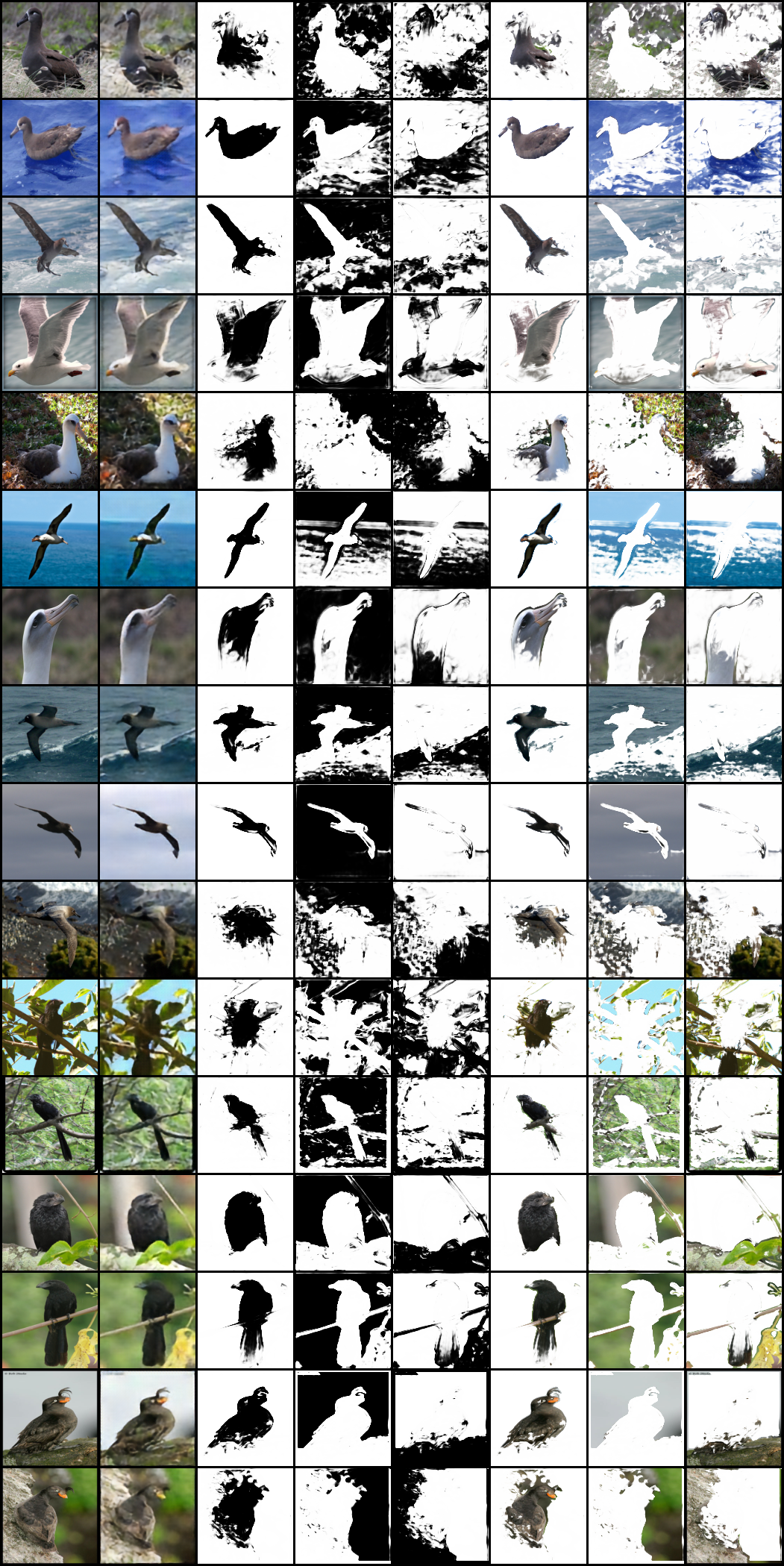}}\hfill
    \resizebox{0.5\linewidth}{!}{\includegraphics{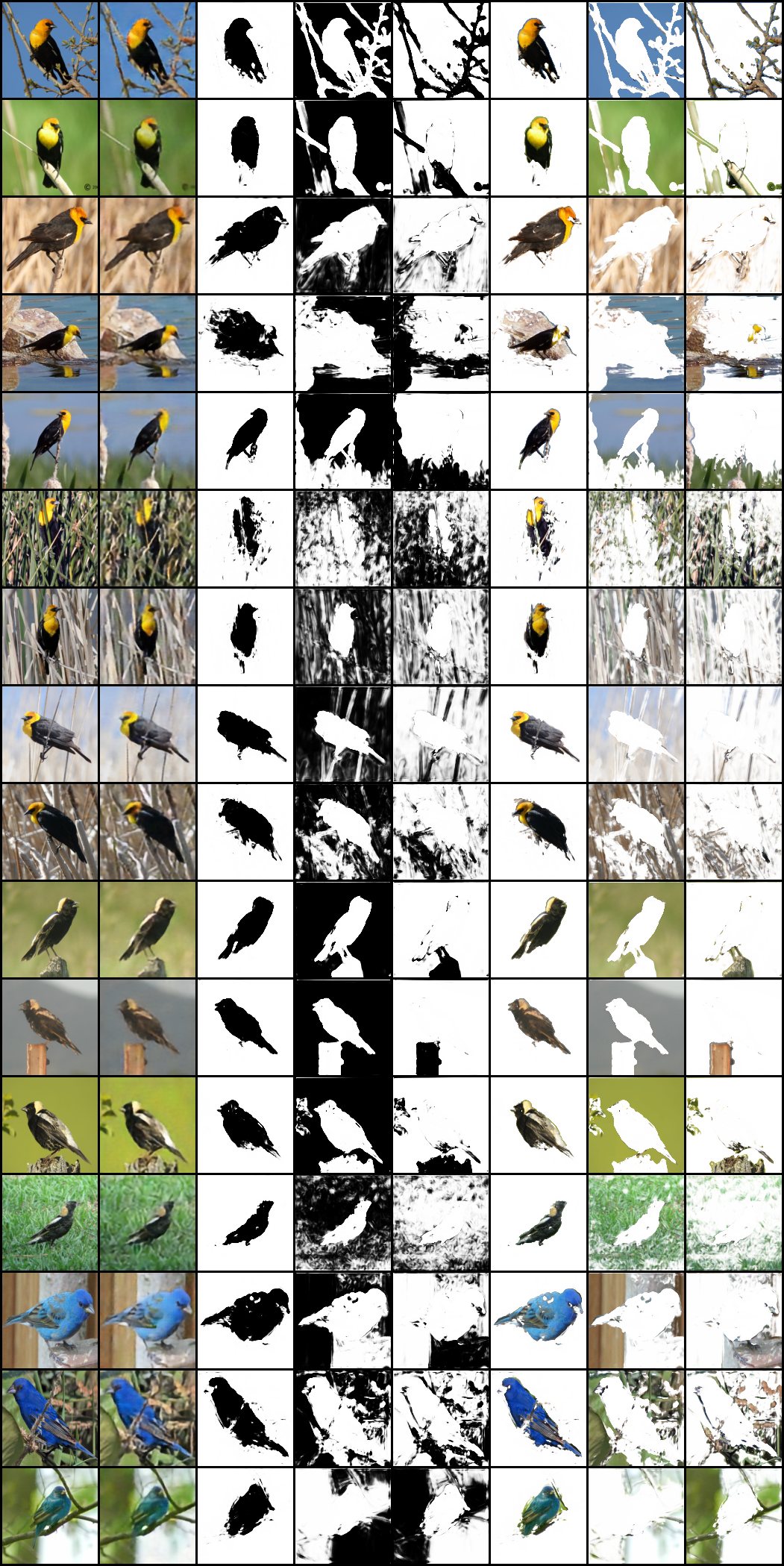}}
    \caption{Unsupervised Foreground Extraction on CUB200 Birds.}
    \label{fig:add:birds}
\end{figure}
\begin{figure}[ht!]
    \centering
    \resizebox{0.5\linewidth}{!}{\includegraphics{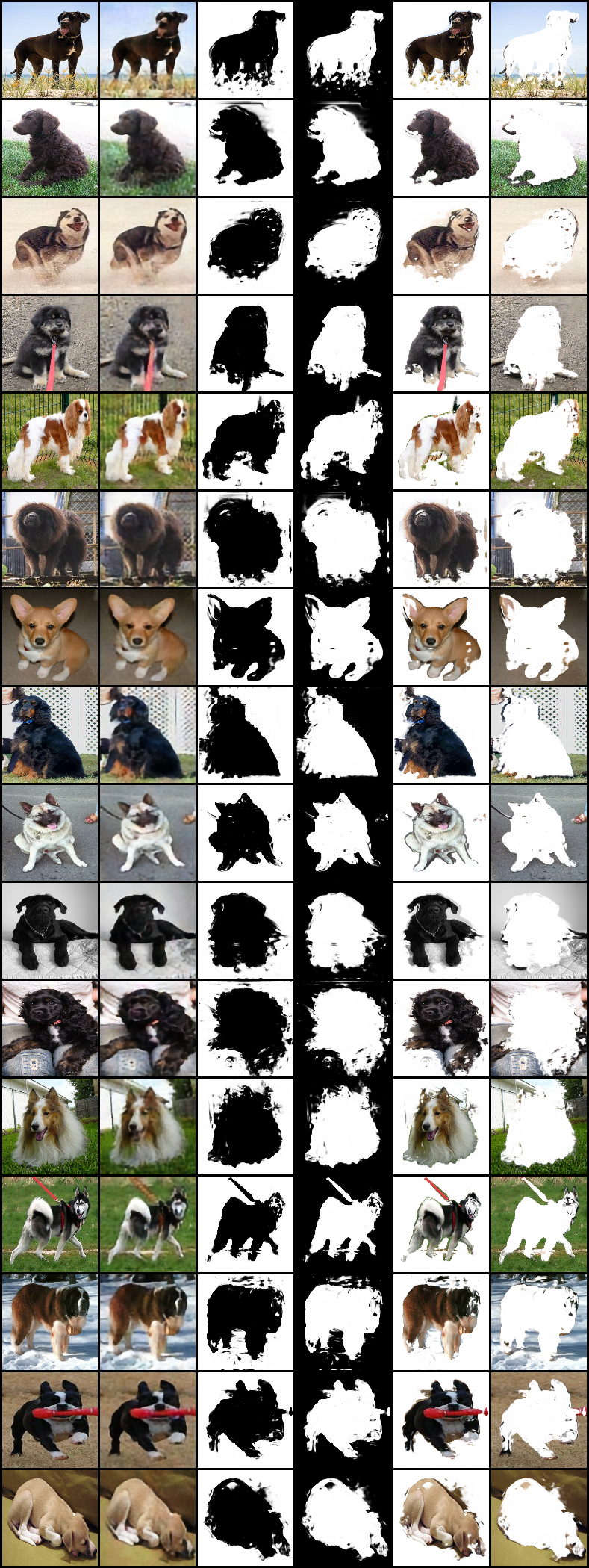}}\hfill
    \resizebox{0.5\linewidth}{!}{\includegraphics{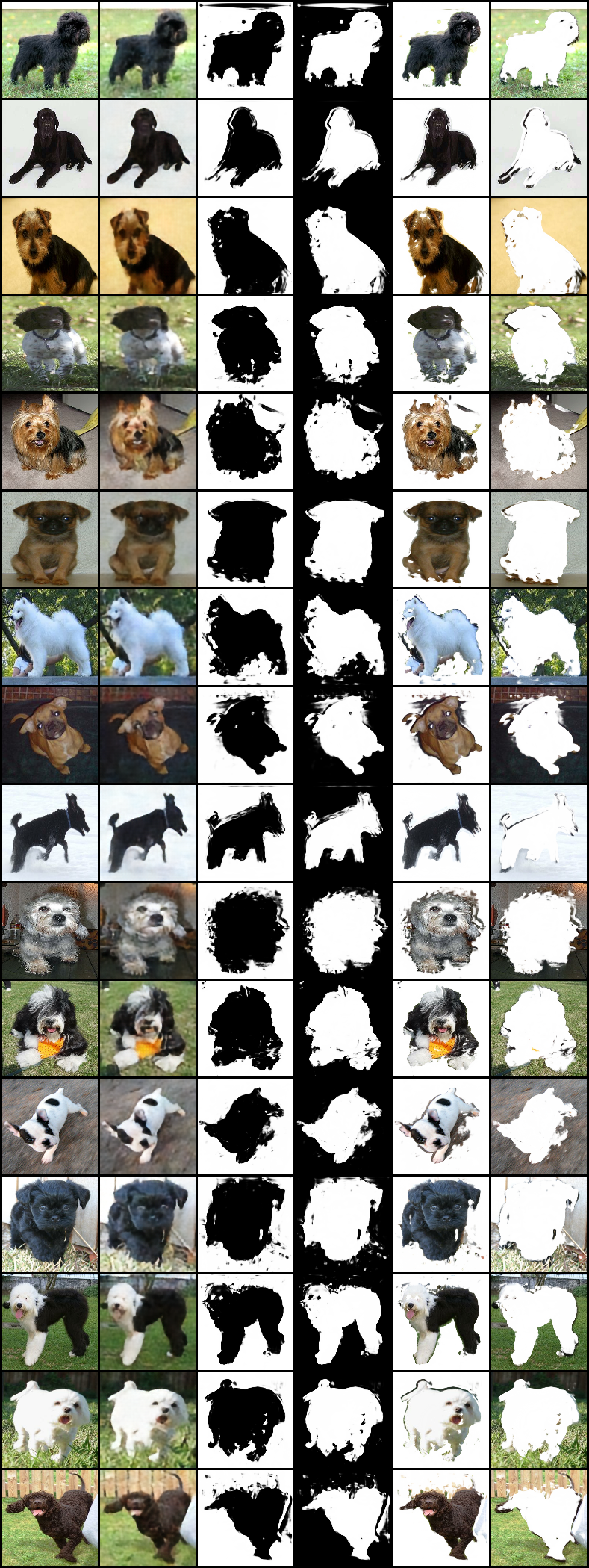}}
    \caption{Unsupervised Foreground Extraction on Stanford Dogs.}
    \label{fig:add:dogs}
\end{figure}
\begin{figure}[ht!]
    \centering
    \resizebox{0.5\linewidth}{!}{\includegraphics{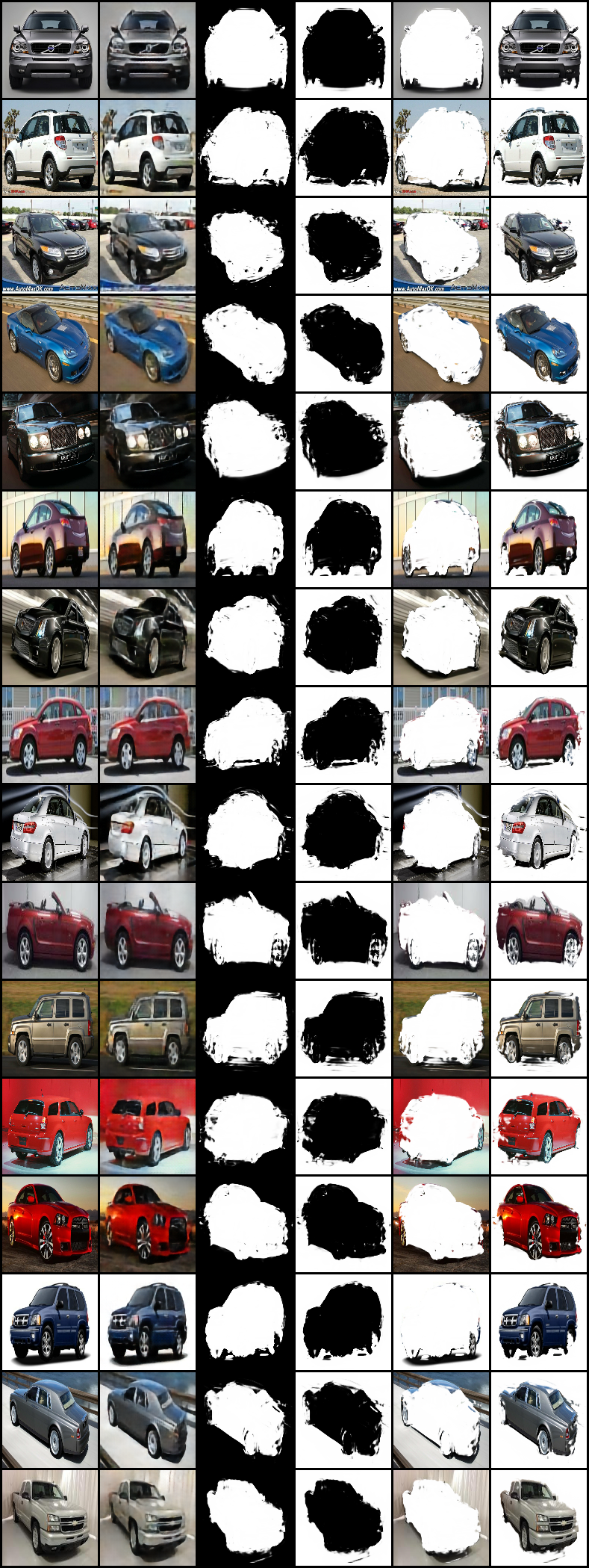}}\hfill
    \resizebox{0.5\linewidth}{!}{\includegraphics{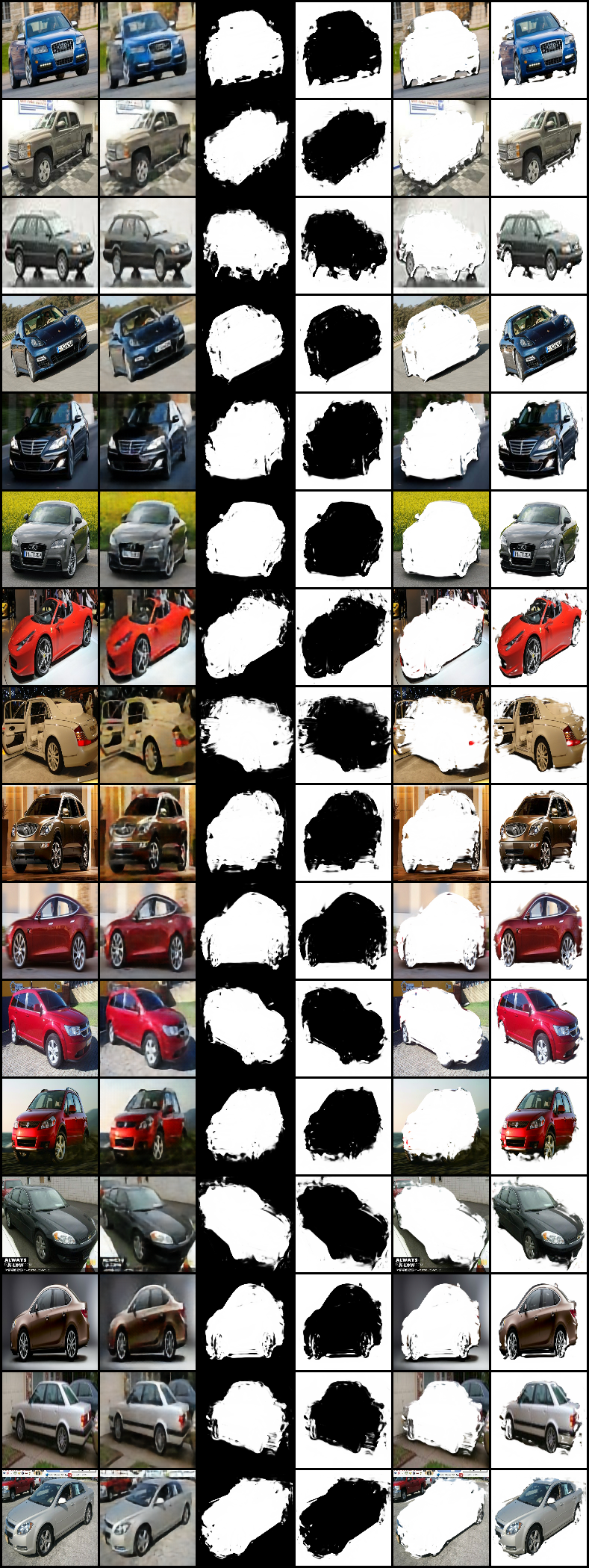}}
    \caption{Unsupervised Foreground Extraction on Stanford Cars.}
    \label{fig:add:cars}
\end{figure}
\begin{figure}[ht!]
    \centering
    \resizebox{0.5\linewidth}{!}{\includegraphics{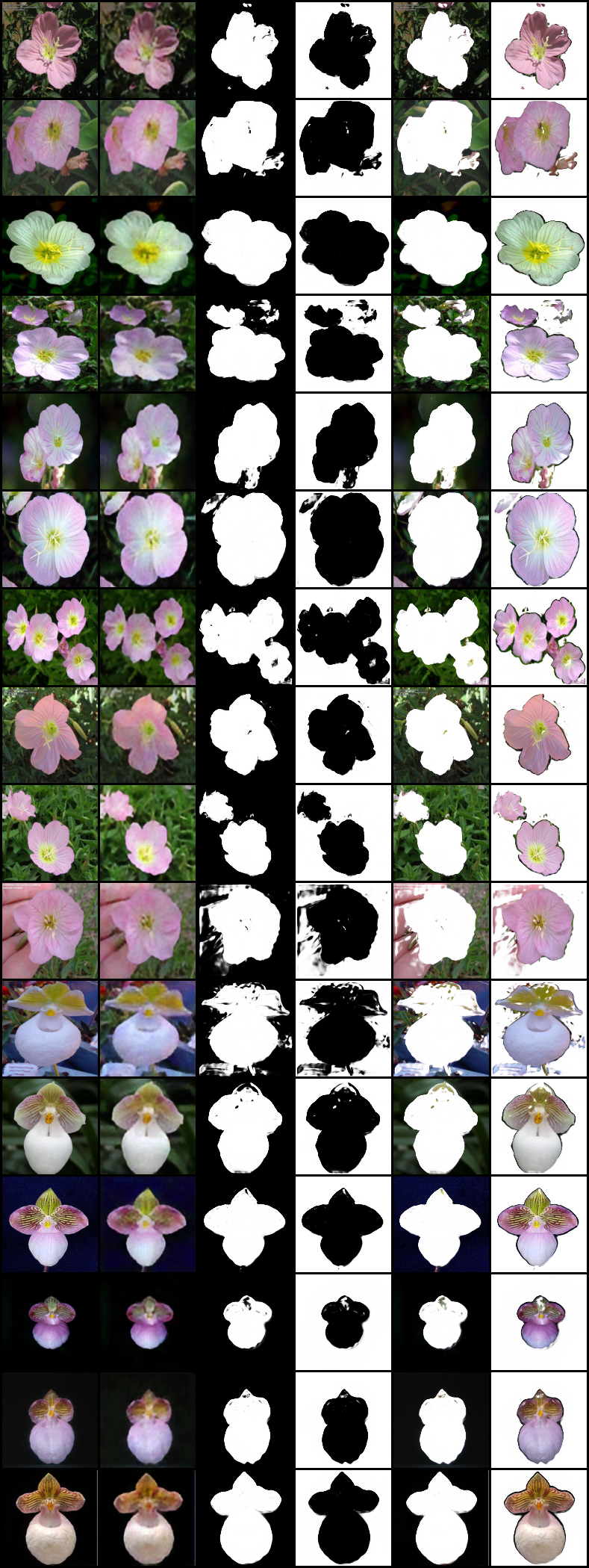}}\hfill
    \resizebox{0.5\linewidth}{!}{\includegraphics{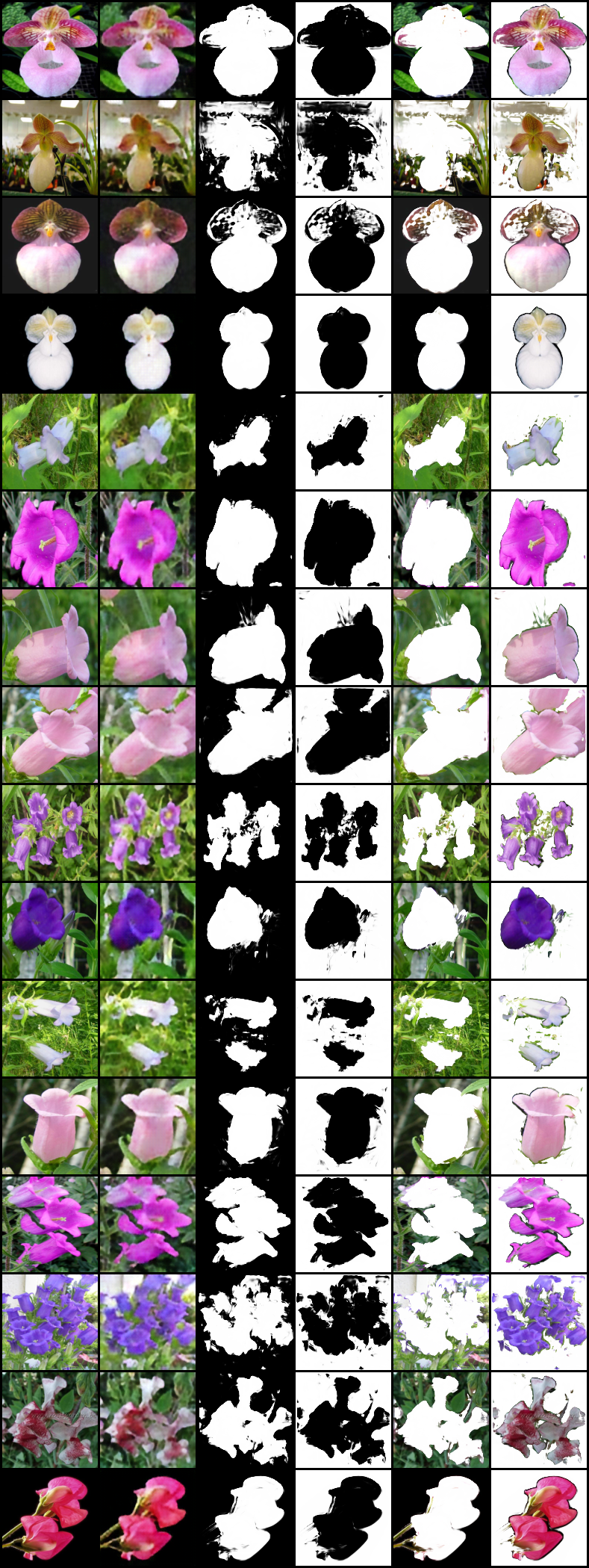}}
    \caption{Unsupervised Foreground Extraction on Caltech Flowers.}
    \label{fig:add:flower}
\end{figure}
\begin{figure}[ht!]
    \centering
    \resizebox{0.75\linewidth}{!}{\includegraphics{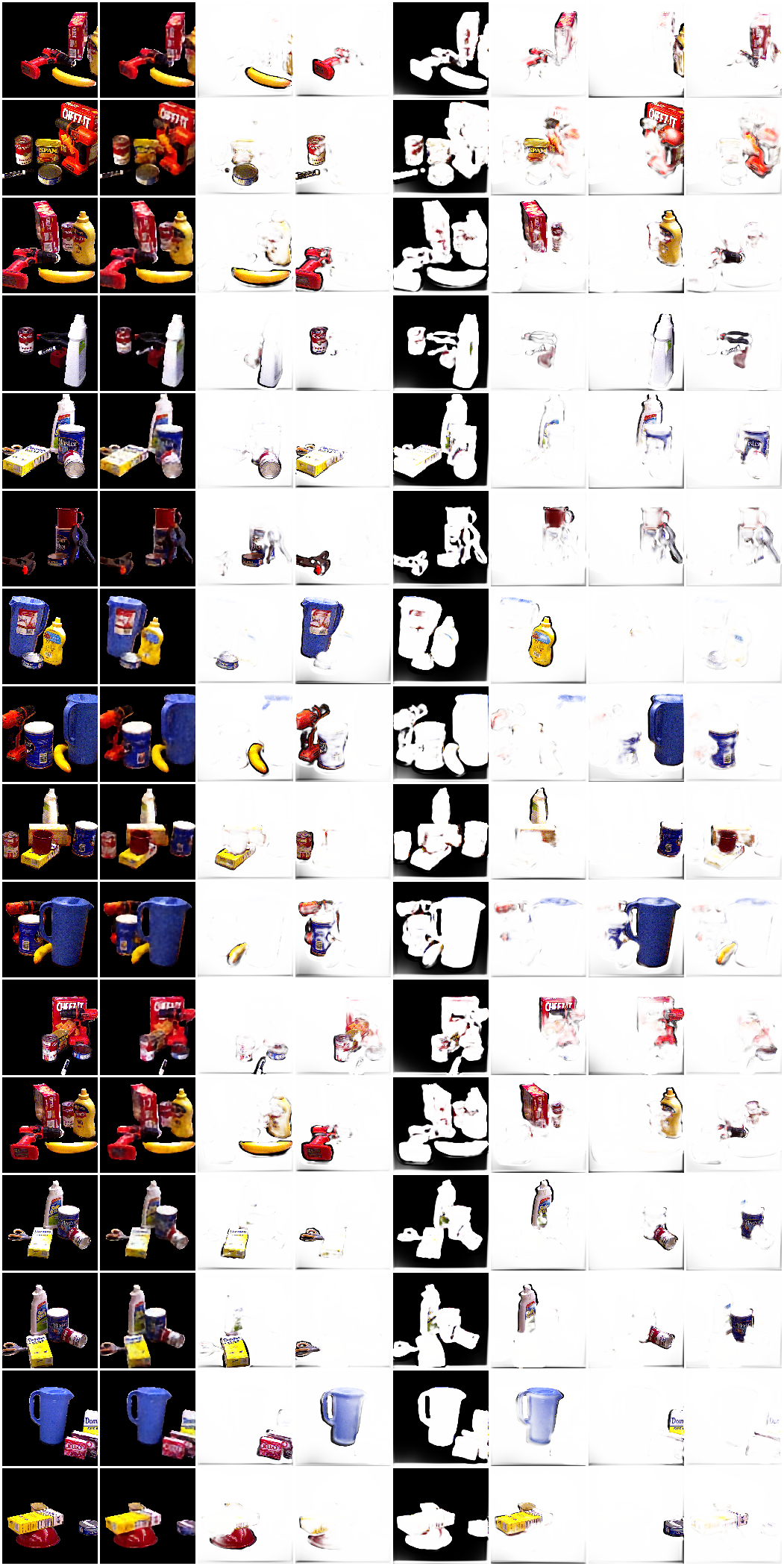}}
    \caption{Unsupervised Multi-Object Segmentation on YCB.}
    \label{fig:add:ycb}
\end{figure}
\begin{figure}[ht!]
    \centering
    \resizebox{0.75\linewidth}{!}{\includegraphics{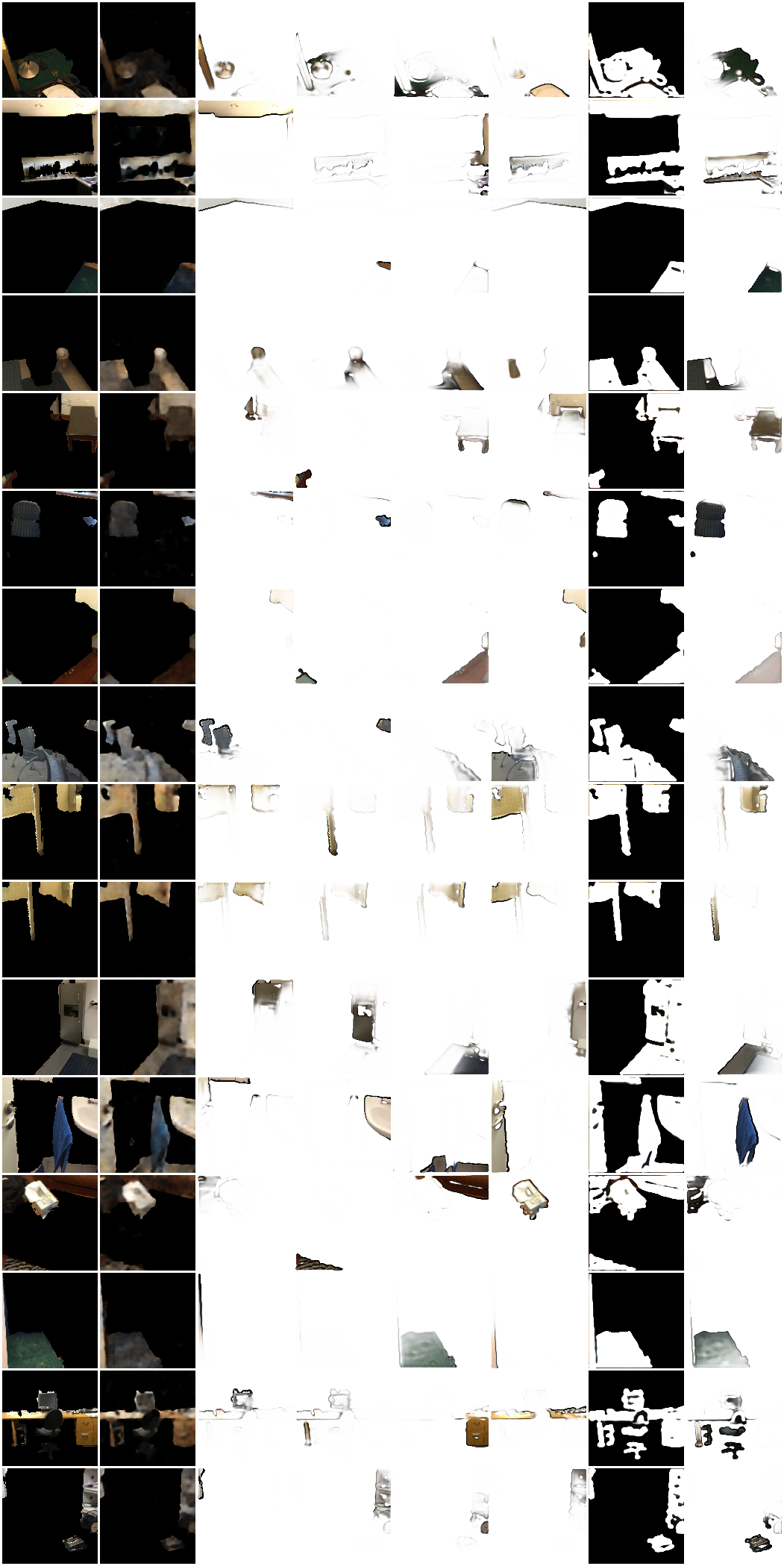}}
    \caption{Unsupervised Multi-Object Segmentation on ScanNet.}
    \label{fig:add:scannet}
\end{figure}
\begin{figure}[ht!]
    \centering
    \resizebox{0.75\linewidth}{!}{\includegraphics{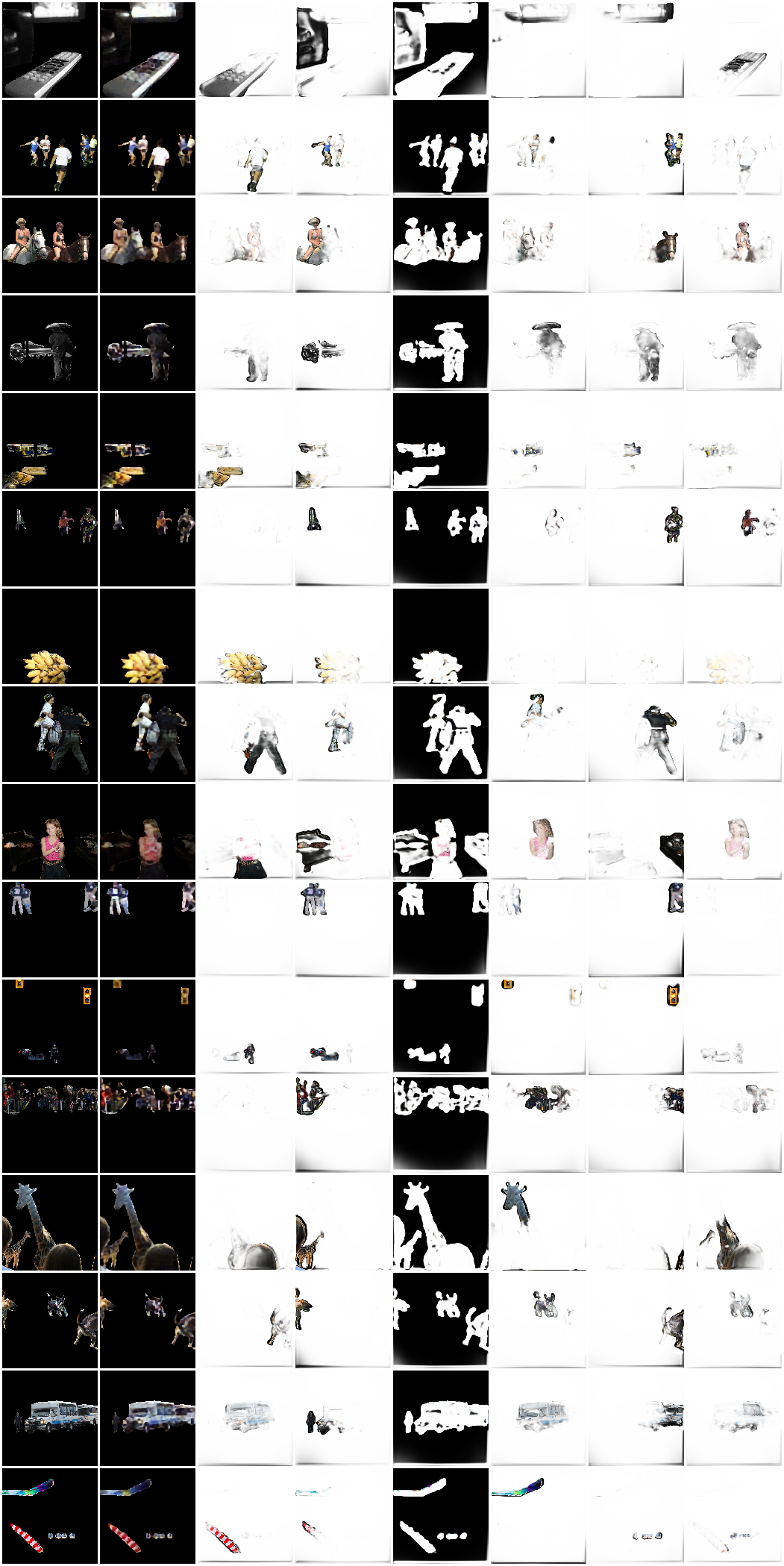}}
    \caption{Unsupervised Multi-Object Segmentation on COCO.}
    \label{fig:add:coco}
\end{figure}
\end{document}